\documentclass[sigconf]{acmart}

\copyrightyear{2021}
\acmYear{2021}
\setcopyright{acmcopyright}
\acmConference[KDD '21] {Proceedings of the 27th ACM SIGKDD Conference on Knowledge Discovery and Data Mining}{August 14--18, 2021}{Virtual Event, Singapore.}
\acmBooktitle{Proceedings of the 27th ACM SIGKDD Conference on Knowledge Discovery and Data Mining (KDD '21), August 14--18, 2021, Virtual Event, Singapore}
\acmPrice{15.00}
\acmISBN{978-1-4503-8332-5/21/08}
\acmDOI{10.1145/3447548.3467403}

\usepackage{graphicx}
\usepackage{hyperref}
\usepackage{color}
\usepackage{amsthm}
\usepackage{bm}
\usepackage{multirow}
\usepackage{amsfonts}

\usepackage[labelformat=simple]{subcaption}

\usepackage{threeparttable}

\begin{document}

\title{Analysis and Applications of Class-wise Robustness in Adversarial Training}

\author[Q. Tian, K. Kuang, K. Jiang, F. Wu, Y. Wang]{
	Qi Tian$^{1}$, Kun Kuang$^{1\dagger}$, Kelu Jiang$^{1}$, Fei Wu$^{1}$, Yisen Wang$^{2\dagger}$
}
\affiliation{
	$^1$ College of Computer Science and Technology, Zhejiang University
}
\affiliation{
	$^2$ Key Lab of Machine Perception (MoE), School of EECS, Peking University
}
\email{
	{tianqics, kunkuang, jiangkelu, wufei}@zju.edu.cn
}
\email{
	{yisen.wang}@pku.edu.cn
}


\begin{abstract}
Adversarial training is one of the most effective approaches to improve model robustness against adversarial examples. 
However, previous works mainly focus on the overall robustness of the model, and the in-depth analysis on the role of each class involved in adversarial training is still missing.
In this paper, we propose to analyze the class-wise robustness in adversarial training. First, we provide a detailed diagnosis of adversarial training on six benchmark datasets, \textit{i.e.}, MNIST, CIFAR-10, CIFAR-100, SVHN, STL-10 and ImageNet.
Surprisingly, we find that there are \textit{remarkable robustness discrepancies among classes}, leading to unbalance/unfair class-wise robustness in the robust models. Furthermore, we keep investigating the relations between classes and find that the unbalanced class-wise robustness is pretty consistent among different attack and defense methods. Moreover, we observe that the stronger attack methods in adversarial learning achieve performance improvement mainly from a more successful attack on the vulnerable classes (\textit{i.e.}, classes with less robustness). Inspired by these interesting findings, we design a simple but effective attack method based on the traditional PGD attack, named Temperature-PGD attack, which proposes to enlarge the robustness disparity among classes with a temperature factor on the confidence distribution of each image. Experiments demonstrate our method can achieve a higher attack rate than the PGD attack.
Furthermore, from the defense perspective, we also make some modifications in the training and inference phase to improve the robustness of the most vulnerable class, so as to mitigate the large difference in class-wise robustness.
We believe our work can contribute to a more comprehensive understanding
of adversarial training as well as rethinking the class-wise properties in robust models.
\end{abstract}



\maketitle

\section{Introduction}
Deep learning has achieved great success in many applications (such as image classification \citep{he2016deep}, video processing \citep{zhang2020comprehensive}, recommender systems \citep{zhang2021cause}).
Unfortunately, the existence of adversarial examples \citep{szegedy2013intriguing} reveals the vulnerability of deep neural networks, which hinders the practical deployment of deep learning models.
Adversarial training (training on adversarial examples) \citep{madry2018towards} has been demonstrated to be one of the most successful defense methods by \citet{athalye2018obfuscated}. 
While it can only obtain moderate robustness even for simple image datasets like CIFAR-10, a comprehensive understanding for adversarial training is critical for further robustness improvement. 

Previously, some works tried to analyze adversarial training from robust optimization \citep{wang2019convergence}, robustness generalization \citep{raghunathan2019adversarial,wu2020adversarial}, training strategy \citep{madry2018towards,zhang2019theoretically,wang2019improving,pang2020boosting,bai2021improving}. However, in these works, they all focus on the averaged robustness over all classes while ignoring the possible difference among different classes. In other fields, there are some works revealing the class-bias learning phenomenon in the standard training (training on natural examples) \citep{tang2020long,wang2019symmetric}, in which they found that some classes (``easy'' classes) are easy to learn and converge faster than other classes (``hard'' classes). Inspired by this, a natural question is then raised here:
\begin{quote}
	\textit{Does each class perform similarly in the adversarially trained models? Or is each class equally vulnerable? If not, how would the class-wise robustness affect the performance of classical attack and defense methods in adversarial learning?}
\end{quote}

\renewcommand{\thefootnote}{\fnsymbol{footnote}}
\footnotetext[2]{Corresponding Authors.}
\renewcommand{\thefootnote}{\arabic{footnote}}

In this paper, we investigate the above questions comprehensively. Specifically, we conduct a series of experiments on several benchmark datasets and find that the class-bias learning phenomenon still exists in adversarial training which is even severe than standard training. For this finding, we have the following questions to explore:
\begin{figure*}[htbp] 
	\centering
	\begin{subfigure}{0.27\textwidth} 
		\centering
		\includegraphics[width=\textwidth]{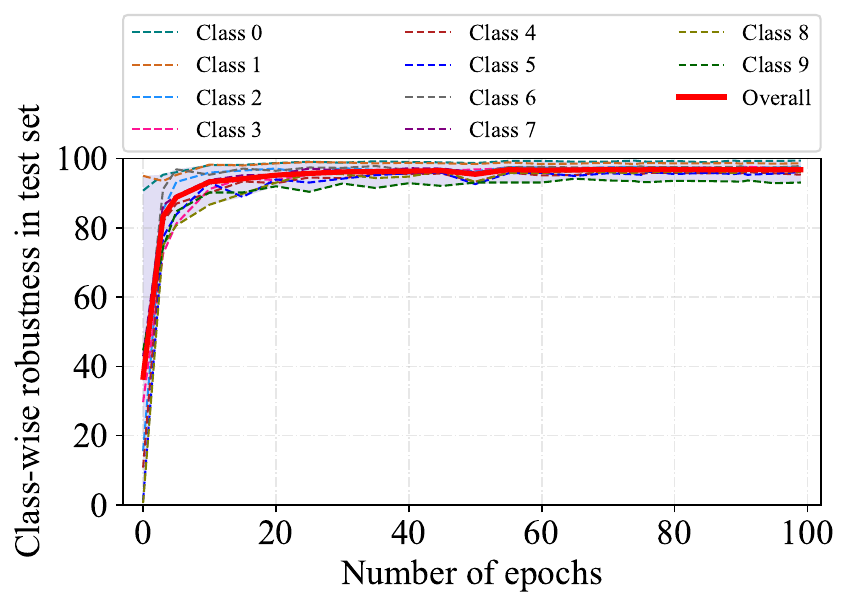}
		\caption{MNIST}
		\label{fig1-MNIST}
	\end{subfigure} \hfill
	\begin{subfigure}{0.27\textwidth}
		\centering
		\includegraphics[width=\textwidth]{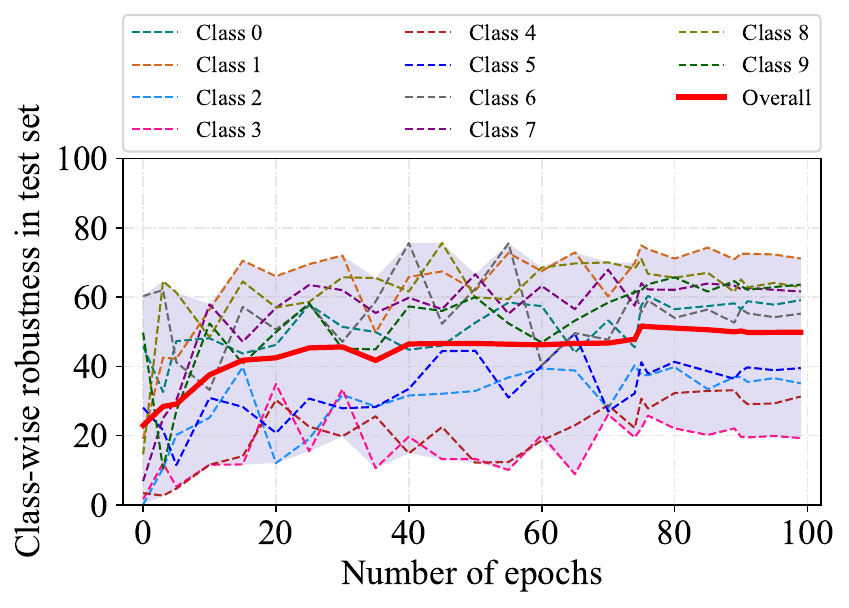}
		\caption{CIFAR-10}
		\label{fig1-CIFAR-10}
	\end{subfigure} \hfill
	\begin{subfigure}{0.27\textwidth}
		\centering
		\includegraphics[width=\textwidth]{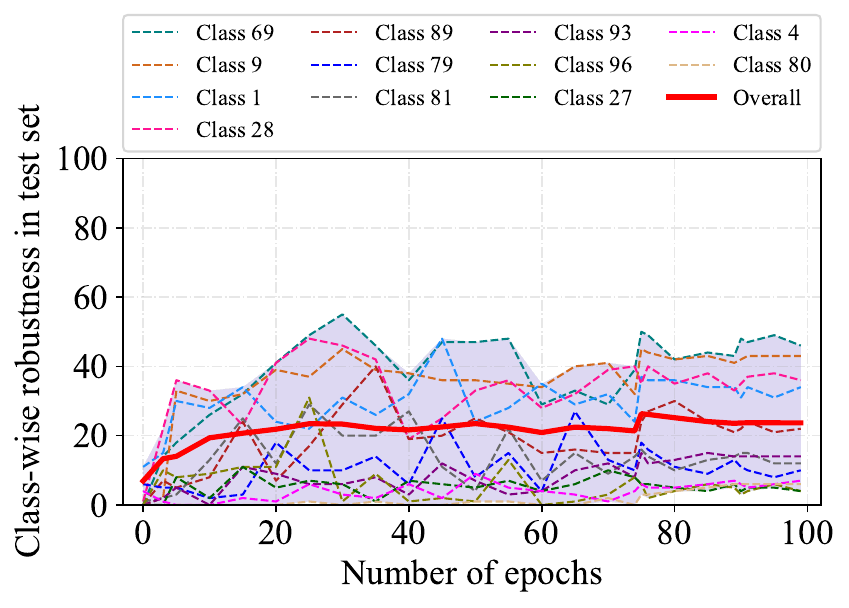}
		\caption{CIFAR-100}
		\label{fig1-CIFAR-100}
	\end{subfigure}
	
	\begin{subfigure}{0.27\textwidth}
		\centering
		\includegraphics[width=\textwidth]{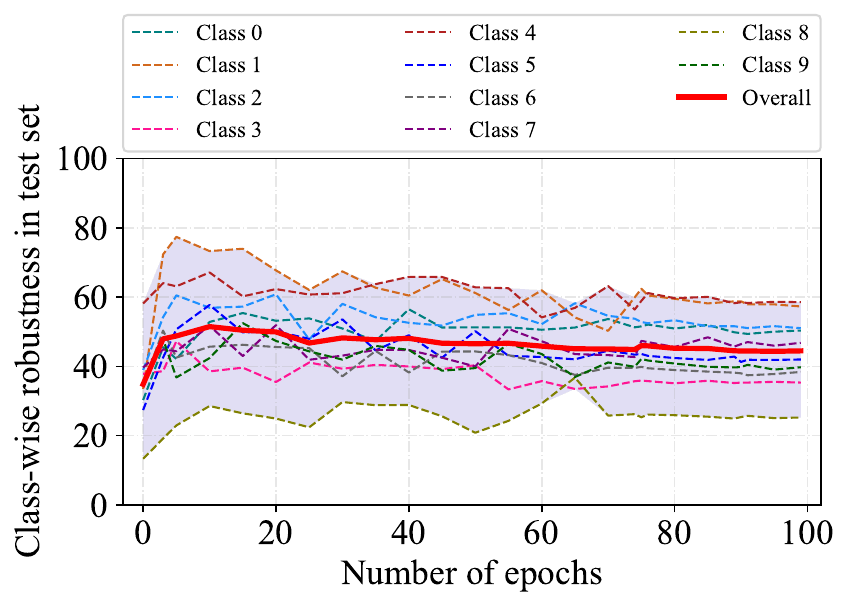}
		\caption{SVHN}
		\label{fig1-SVHN}
	\end{subfigure} \hfill
	\begin{subfigure}{0.27\textwidth}
		\centering
		\includegraphics[width=\textwidth]{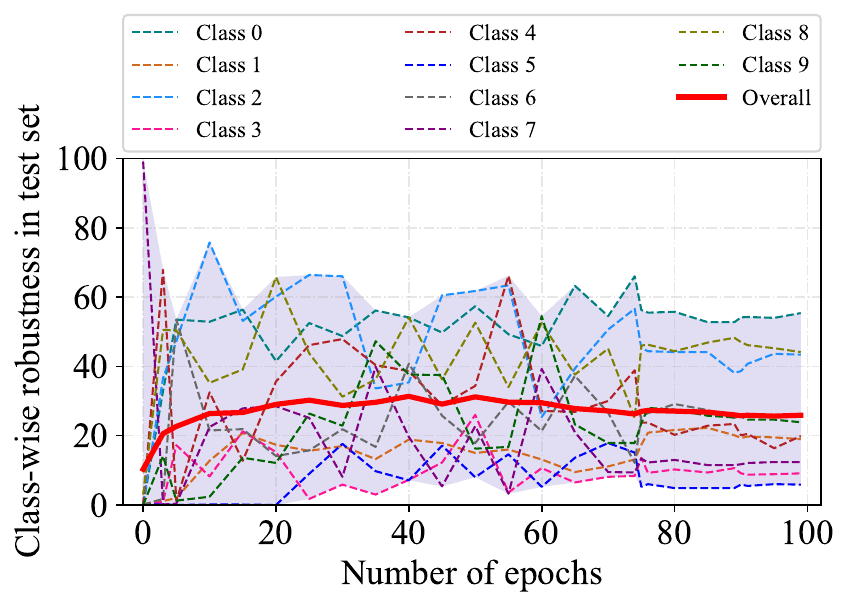}
		\caption{STL-10}
		\label{fig1-STL-10}
	\end{subfigure} \hfill
	\begin{subfigure}{0.27\textwidth}
		\centering
		\includegraphics[width=\textwidth]{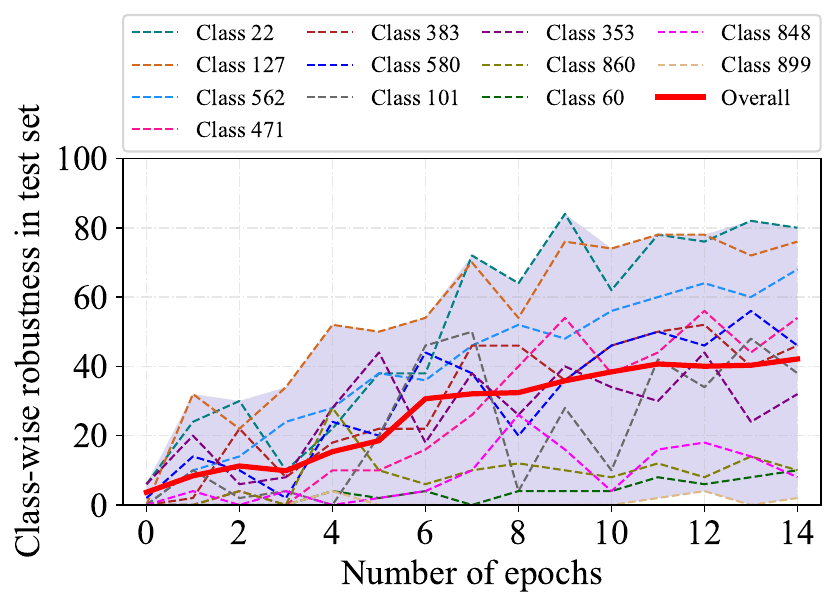}
		\caption{ImageNet}
		\label{fig1-ImageNet}
	\end{subfigure}
	\caption{Class-wise robustness at different epochs in the test set}
	\label{fig1}
\end{figure*}

\begin{itemize}
	\item[1)] What is the relation between the unbalanced robustness and the properties of the dataset itself?
	\item[2)] Can we use the class-wise properties to further enlarge the differences among classes?
	\item[3)] Are there any ways to improve the robustness of vulnerable classes so as to obtain a more balance/fairer robust model?
\end{itemize}
We conduct extensive analysis on the obtained robust models and summarize the following contributions:
\begin{itemize}
	\item \textbf{Analysis on class-wise robustness}
	\begin{enumerate}
		\item[1)] We systematically investigate the relation between different classes and find classes in each dataset can be divided into several groups, and intra-group classes are easily affected by each other.
		\item[2)] The relative robustness between each class is pretty consistent among different attack or defense methods, which indicates that the dataset itself plays an important role in the class-wise robustness.
	\end{enumerate}
	\item \textbf{Applications for stronger attack}
	\begin{enumerate}
		\item[1)] We make full use of the properties of the vulnerable classes to propose an attack that can effectively reduce the robustness of these classes, thereby increasing the disparity among classes.
	\end{enumerate}
	\item \textbf{Applications for stronger defenses}
	\begin{enumerate}
		\item[1)] Training phase: Since the above group-based relation is commonly observed in the dataset, we propose a method that can effectively use this relation to adjust the robustness of the most vulnerable class.		
		\item[2)] Inference phase: We find that the background of the images may be a potential factor for different classes to be easily flipped by each other. Our experiments show that the robustness of the most vulnerable class can be improved by simply changing the background.
	\end{enumerate}
\end{itemize}

\section{Related Work} \label{sec2}

\; \; \textbf{Class-wise analysis.}\, 
Class-wise properties are widely studied in the deep learning community, such as long-tailed data \citep{tang2020long} and noisy label \citep{wang2019symmetric}.
The datasets for these specific tasks are significantly different in each class.
\textit{i.e.}, in long-tailed data task, the tail-class (with few training data) usually achieves lower accuracy since it cannot be sufficiently trained. In the asymmetric noisy label task, classes with more label noise usually have lower accuracy.
However, in the adversarial community, few people pay attention to class-wise properties because all benchmark datasets seem to be class-balanced. 
Recently, we notice two parallel and independent works \citep{nanda2021fairness,benz2020robustness} also point out the performance disparity in robust models,
but none of them explore the relation between class-wise robustness and the properties of the dataset itself, and our work takes the first step to investigate this problem.
\begin{figure*}[htbp]
	\centering
	
	\begin{subfigure}{0.27\textwidth}
		\centering
		\includegraphics[width=\textwidth]{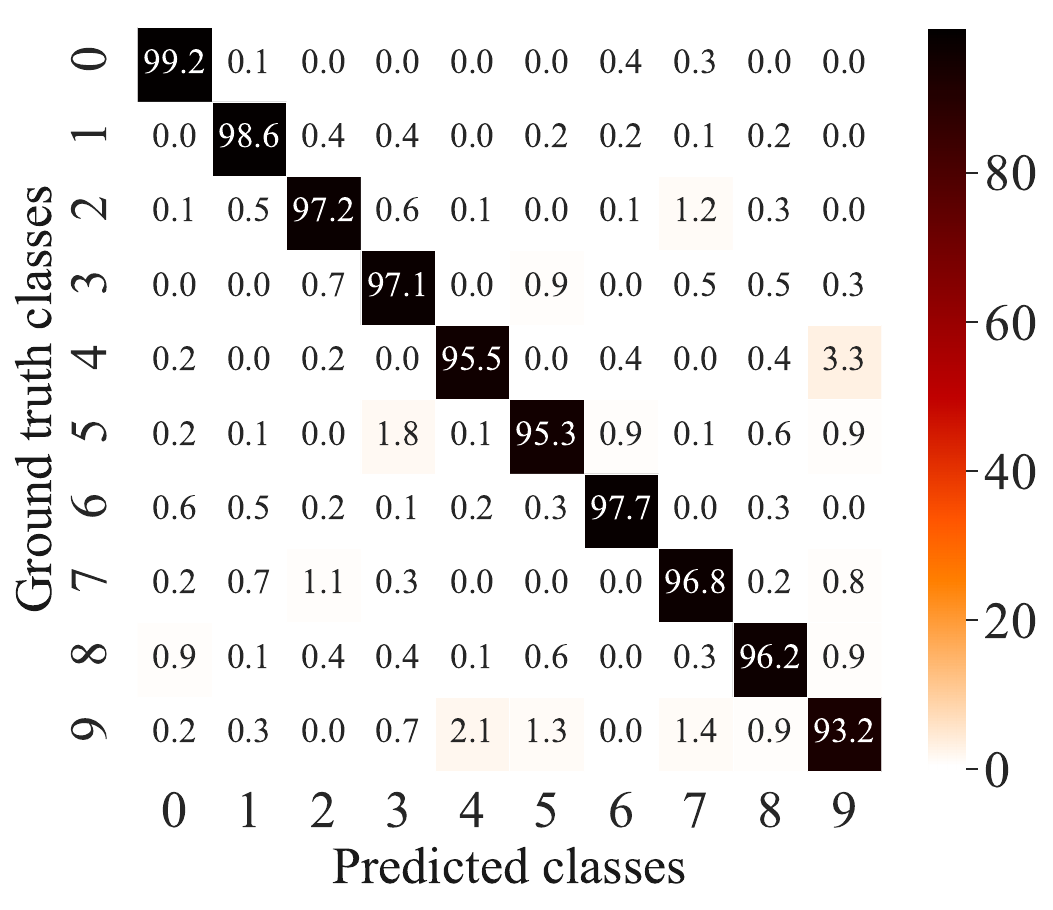}
		\caption{MNIST}
		\label{fig2-MNIST}
	\end{subfigure} \hfill
	\begin{subfigure}{0.27\textwidth}
		\centering
		\includegraphics[width=\textwidth]{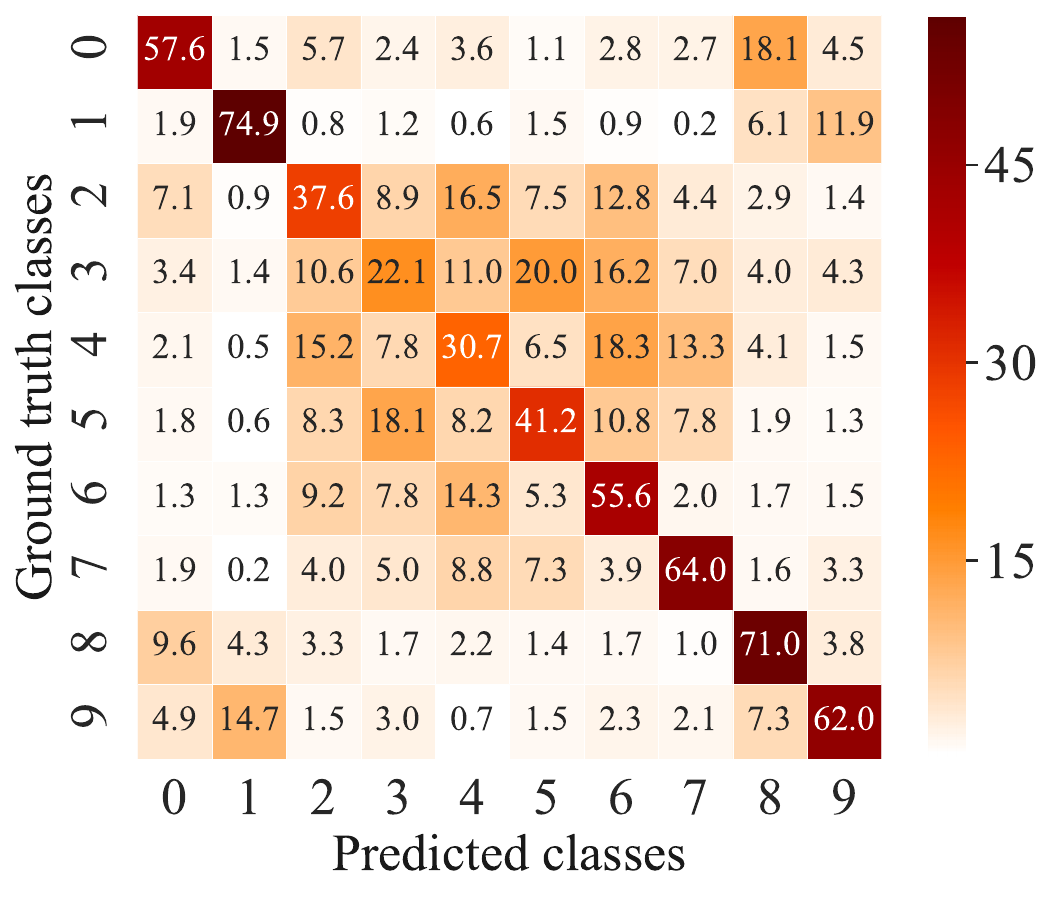}
		\caption{CIFAR-10}
		\label{fig2-CIFAR-10}
	\end{subfigure} \hfill
	\begin{subfigure}{0.27\textwidth}
		\centering
		\includegraphics[width=\textwidth]{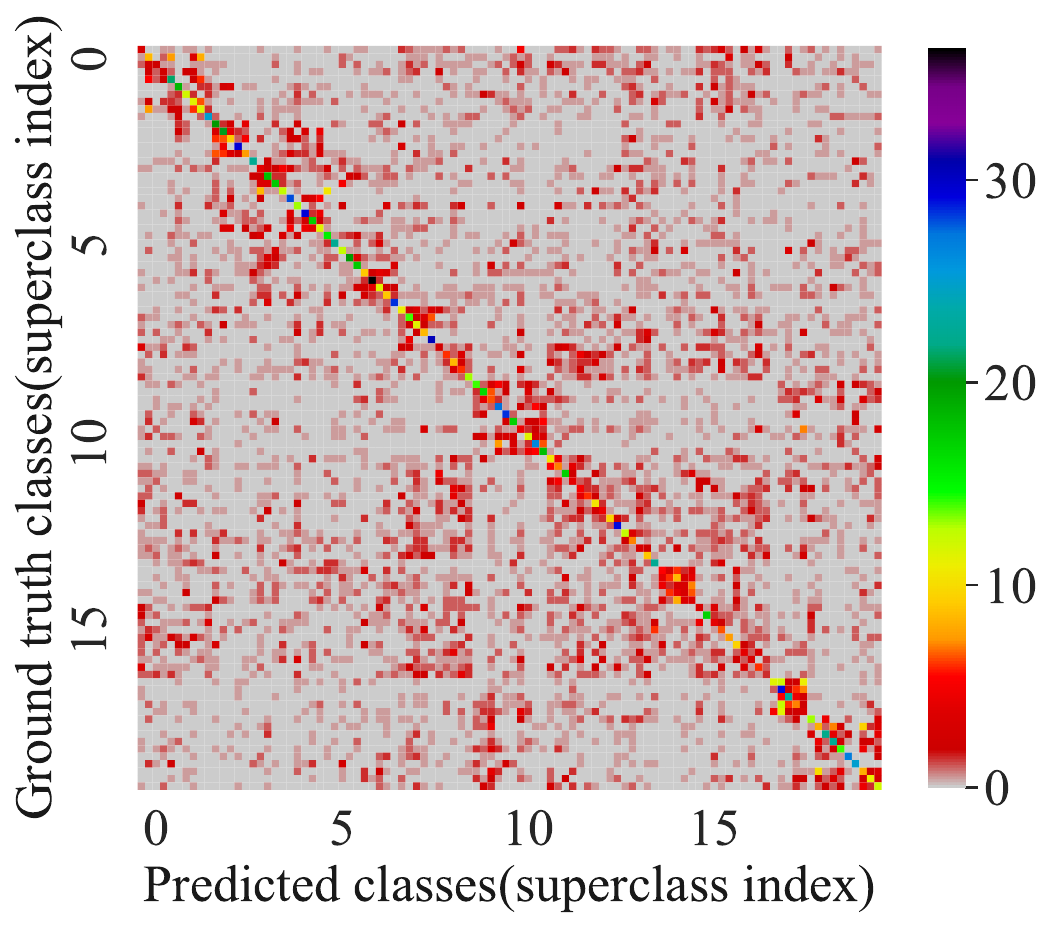}
		\caption{CIFAR-100}
		\label{fig2-CIFAR-100}
	\end{subfigure}
	
	\begin{subfigure}{0.27\textwidth}
		\centering
		\includegraphics[width=\textwidth]{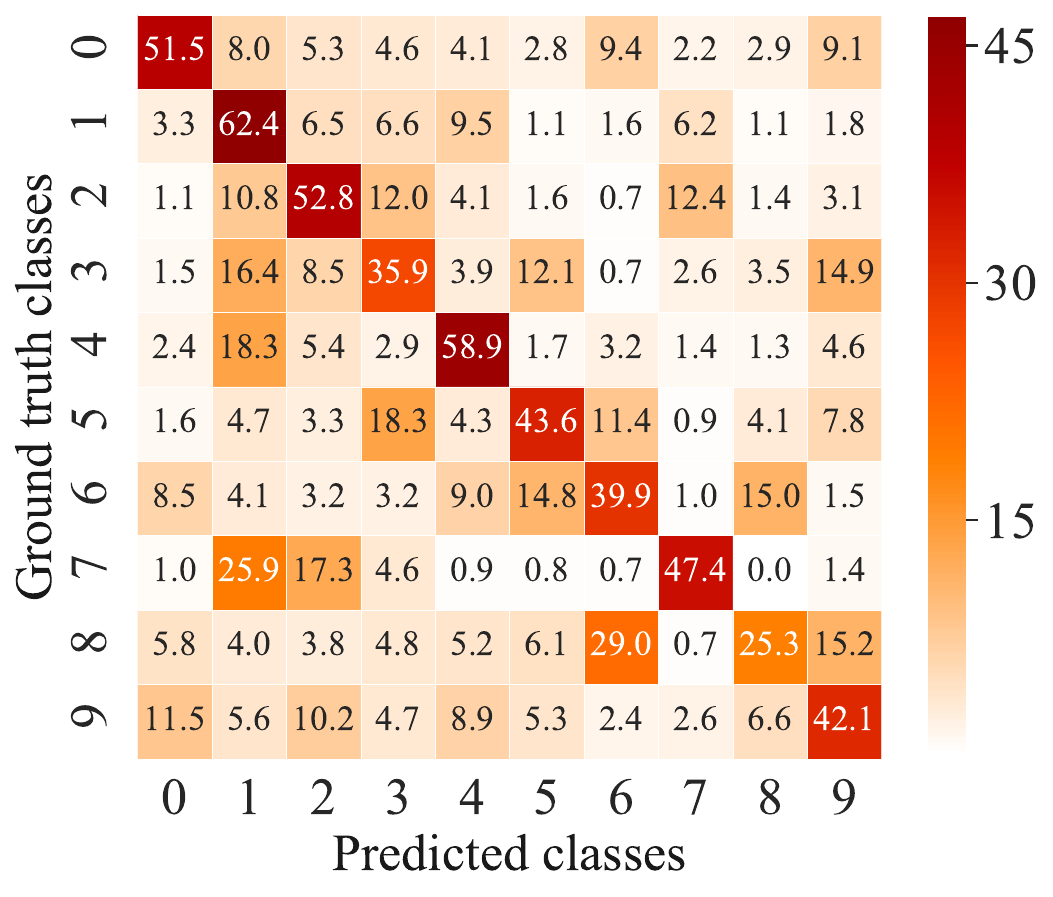}
		\caption{SVHN}
		\label{fig2-SVHN}
	\end{subfigure} \hfill
	\begin{subfigure}{0.27\textwidth}
		\centering
		\includegraphics[width=\textwidth]{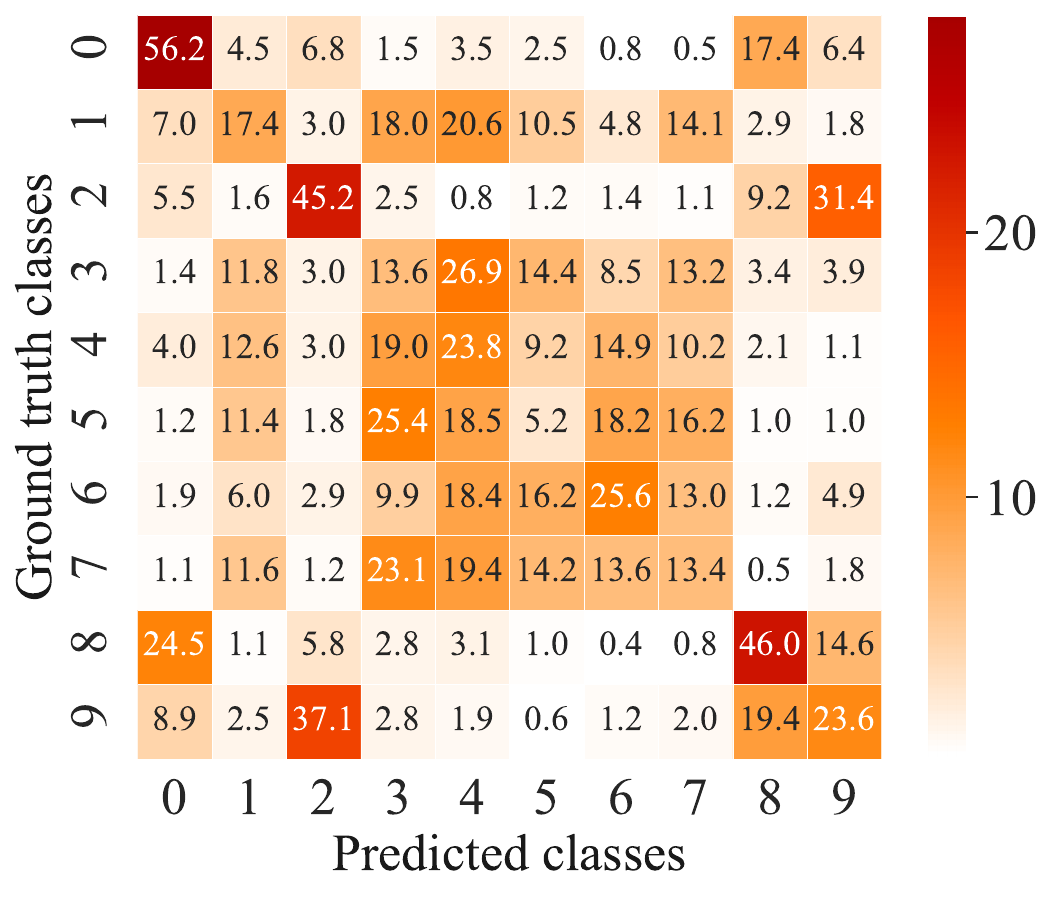}
		\caption{STL-10}
		\label{fig2-STL-10}
	\end{subfigure} \hfill
	\begin{subfigure}{0.27\textwidth}
		\centering
		\includegraphics[width=\textwidth]{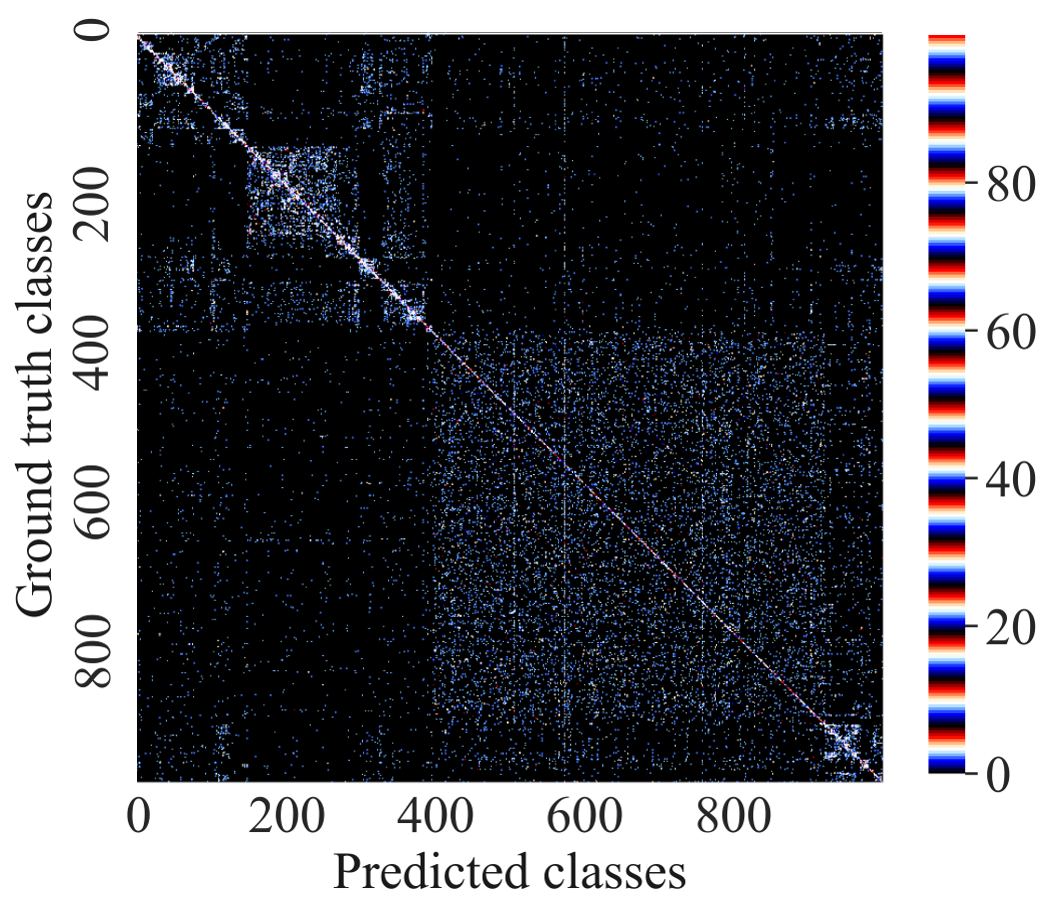}
		\caption{ImageNet}
		\label{fig2-ImageNet}
	\end{subfigure}
	\caption{Confusion matrix of robustness in the test set}
	\label{fig2}
\end{figure*}

\textbf{Attack.}\,
Adversarial attacks are used to craft adversarial examples by adding small and human imperceptible adversarial perturbations to natural examples, which mainly include white-box attacks and black-box attacks. 
In white-box settings, the attackers know the parameters of the defender model and generate adversarial noise by maximizing the loss function (\textit{e.g.}, Fast Gradient Sign Method (FGSM) \citep{goodfellow2014explaining} and Projected Gradient Descent (PGD) \citep{madry2018towards} attack maximize cross-entropy loss, while Carlini-Wagner (C\&W) \citep{carlini2017towards} attack maximize hinge loss).
In black-box settings, there are transfer-based and query-based attacks. The former attacks a substitute model and the generated noise can transfer to the target model \citep{wu2019skip,wang2020unified}. The latter crafts adversarial examples by querying the output of the target model \citep{li2019nattack,bai2020improving}.
In this paper, we analyze the class-wise robustness performance of different attacks and propose an attack to illustrate that the unbalanced robustness can be enlarged by carefully using the information of vulnerable classes.

\textbf{Defense.}\,
Adversarial training \citep{madry2018towards} is known as the most effective and standard
way to against adversarial examples. A range of methods have been proposed to improve adversarial training, including modifying regularization term \citep{zhang2019theoretically,wang2019improving,pang2020boosting}, adding
unlabeled data \citep{carmon2019unlabeled} and data augmentation \citep{song2019robust}. Since adversarial training is more time-consuming than standard training,
\citet{wong2019fast} propose some solutions to accelerate model training.
On the other hand, 
some researchers \citep{raff2019barrage,pang2019mixup,bai2019hilbert} try to improve model robustness by preprocessing the image in the inference phase, and these methods are usually complementary to adversarial training.
However, none of these methods consider the difference in class-wise robustness, and we have proposed some methods that can improve the robustness of the most vulnerable class so as to obtain a fairer output.

\section{Preliminary}
In this section, we first introduce the formula and notations in adversarial training, then give several definitions about robust/non-robust example and robust/vulnerable/confound class used through this paper. 

\textbf{Vanilla adversarial training.}\,
\citet{madry2018towards} formalize the adversarial training as a min-max optimization problem. Given a DNN classifier $h_{\bm{\theta}}$ with parameters $\bm{\theta}$, a correctly classified natural example $\bm{x}$ with class label $y$, cross-entropy loss $\ell(\cdot)$ and an adversarial example $\bm{x}^\prime$ can be generated by perturbing $\bm{x}$, then the objective of adversarial training is:
\begin{align}
	\label{eq1}
	\textstyle
	\min\limits_{\bm{\theta}} \frac{1}{n} \sum_{i=1}^{n} \max\limits _{\left\|{\bm{x}}_{i}^{\prime}-{\bm{x}}_{i}\right\|_{p} \leq \epsilon} \ell\left(h_{\bm{\theta}}\left({\bm{x}}_{i}^{\prime}\right), y_{i}\right),
\end{align}
where the inner maximization applies the Projected Gradient Descent (${\rm PGD}$) attack to craft adversarial examples, and the outer minimization uses these examples as augmented data to train the model.
Since the adversarial perturbation should not be observed by humans, these noises are bounded by $L_p$-norm $\left\|{\bm{x}}_{i}^{\prime}-{\bm{x}}_{i}\right\|_{p} \leq \epsilon$. 

\textbf{TRADRS.}\,
Another popular adversarial training method (TRADRS \citep{zhang2019theoretically}) is to add a regularization term to the cross-entropy loss:
\begin{align}
	\label{eq2}
	\textstyle
	\min\limits_{\bm{\theta}} \frac{1}{n} \sum_{i=1}^{n} \ell\left(h_{\bm{\theta}}\left({\bm{x}}_{i}\right), y_{i}\right) + \beta \max\limits _{\left\|{\bm{x}}_{i}^{\prime}-{\bm{x}}_{i}\right\|_{p} \leq \epsilon} \mathcal{K} \left(h_{\bm{\theta}}\left({\bm{x}}_{i}\right), h_{\bm{\theta}}\left({\bm{x}}_{i}^{\prime}\right)\right),
\end{align}
where $\mathcal{K}(\cdot)$ represents Kullback-Leibler divergence and $\beta$ can adjust the relative performance between natural and robust accuracy.

In addition, we define some concepts for the convenience of the following expressions.

\begin{definition}
	(\textbf{Robust Example}) Given a natural example $\bm{x}$  with ground truth class $y$ and a DNN classifier $h_{\bm{\theta}}$ with parameters $\bm{\theta}$, if this example does not exist adversarial counterpart in bounded $\epsilon$-ball $\left\|{\bm{x}}^{\prime}-{\bm{x}}\right\|_{p} \leq \epsilon$: $h_{\bm{\theta}}({\bm{x}}^{\prime})\equiv y$, the example $\bm{x}$ is defined as a robust example.
\end{definition}

\begin{definition}
	\label{def_2}
	(\textbf{Non-Robust Example and Confound Class}) Given a natural example $\bm{x}$ with ground truth class $y$ and a DNN classifier $h_{\bm{\theta}}$ with parameters $\bm{\theta}$, if the prediction of the model is $y^{\prime}$ after adding a bounded $\epsilon$-ball $\left\|{\bm{x}}^{\prime}-{\bm{x}}\right\|_{p} \leq \epsilon$ : $h_{\bm{\theta}}({\bm{x}}^{\prime})=y^{\prime}\neq y$, the example $\bm{x}$ is defined as a non-robust example and $y^{\prime}$ is defined as confound class of example $\bm{x}$.
\end{definition}
\begin{table*}[h]
	\scriptsize
	\centering
	\caption{Adversarial robustness (\%) (under popular attacks) on CIFAR-10.}
	\label{table4}
	\begin{threeparttable}
		\begin{tabular}{ccccccccccccccc}
			\toprule
			Defenses(Attacks)       & Tot.  &  0    &  1    &  2    & 3     & 4     & 5     & 6     & 7     & 8     & 9    & CV & MCD \\ \midrule
			Madry(FGSM)   & 65.5&73.7&\underline{81.2}\tnote{1}&51.9&\textbf{41.5}\tnote{2}&54.2&49.4&73.9&72.5&78.5&78.5&191.9&39.7 \\ 
			TRADES(FGSM)  & 66.9&77.5&\underline{85.9}&49.7&\textbf{41.9}&55.8&52.8&73.0&76.8&80.7&75.3&211.2&44.0 \\ 
			MART(FGSM)    & 67.4&73.7&\underline{84.9}&54.5&\textbf{45.7}&50.1&51.6&76.9&75.2&83.9&77.7&206.1&39.2 \\ 
			HE(FGSM)     & 68.4&71.9&\underline{84.6}&52.0&\textbf{42.2}&57.0&57.9&76.5&77.8&83.4&80.9&200.5&42.3 \\ \midrule
			Madry($\rm CW_{\infty}$)   &57.1&67.5&\underline{79.5}&43.0&\textbf{37.7}&41.5&41.0&57.5&60.0&71.5&72.0&212.5&41.8 \\ 
			TRADES($\rm CW_{\infty}$)  &59.4&69.5&\underline{85.5}&39.0&\textbf{38.5}&43.0&46.5&57.0&67.0&77.5&70.5&258.5&47.0 \\ 
			MART($\rm CW_{\infty}$)    &58.8&65.5&\underline{80.5}&43.0&\textbf{39.5}&41.0&41.0&63.0&67.5&76.5&71.0&232.7&41.0 \\ 
			HE($\rm CW_{\infty}$)     &63.6&71.2&\underline{87.5}&47.1&\textbf{44.4}&49.8&50.1&61.4&71.2&81.5&72.7&210.8&43.1 \\ \midrule
			Madry(${\rm PGD}$) &52.1&63.8&\underline{71.6}&39.1&\textbf{25.3}&36.7&38.6&57.4&59.5&63.1&66.8&224.3&46.3 \\ 
			TRADES(${\rm PGD}$)
			&56.3&67.8&\underline{80.6}&37.8&\textbf{29.4}&40.6&43.9&59.3&66.9&71.8&65.6&263.6&51.1 \\ 
			MART(${\rm PGD}$)  &58.2&64.5&\underline{78.0}&45.1&\textbf{35.4}&37.7&43.5&65.3&67.5&76.3&69.5&235.1&42.6 \\ 
			HE(${\rm PGD}$)   &60.7&64.9&\underline{79.3}&41.0&\textbf{34.5}&47.9&51.5&67.6&70.5&76.9&73.2&224.8&44.8 \\ \midrule
			Madry(Transfer-based attack)   &80.2&84.5&\underline{87.7}&71.0&\textbf{68.3}&78.9&69.2&86.3&82.9&87.6&86.4&56.1&19.4 \\ 
			TRADES(Transfer-based attack)  &82.0&87.7&\underline{92.3}&70.9&\textbf{68.0}&78.2&70.0&87.8&87.6&90.8&86.8&78.1&24.2 \\ 
			MART(Transfer-based attack)    &82.9&87.4&\underline{94.7}&74.0&\textbf{66.7}&76.0&68.8&89.9&88.0&93.7&90.0&99.2&28.0 \\ 
			HE(Transfer-based attack)     &84.5&90.1&\underline{95.9}&75.6&\textbf{60.8}&77.4&76.7&91.1&92.1&93.4&92.2&115.3&35.1 \\ \midrule
			Madry($\mathcal{N}$ atacck) &56.1&67.5&\underline{77.7}&43.7&\textbf{31.4}&42.7&49.0&53.7&60.1&64.4&71.1&190.5&46.3 \\
			TRADES($\mathcal{N}$ atacck) &64.4&73.1&\underline{87.4}&46.4&\textbf{44.4}&49.1&61.7&56.9&71.6&79.5&74.1&200.0&43.0 \\
			MART($\mathcal{N}$ atacck)  &67.5&72.3&\underline{83.4}&55.3&\textbf{49.0}&54.1&61.2&67.1&72.9&82.3&77.6&133.6&34.4 \\ 
			HE($\mathcal{N}$ atacck)   &69.7&75.9&\underline{88.3}&52.7&\textbf{44.7}&65.4&62.6&70.1&76.0&84.5&77.5&168.6&43.5 \\ 
			\bottomrule
		\end{tabular}
		\begin{tablenotes}
			\scriptsize
			\item[1] The underscore indicates the most robust class.
			\item[2] The bold indicates the most vulnerable class.
		\end{tablenotes}
	\end{threeparttable}
\end{table*}

\begin{table}[htbp]
	\centering
	\caption{Superclasses in CIFAR-10 and STL-10.}
	\label{table0}
	\footnotesize
	\begin{tabular}{c|c|lccclclcccl}
		\hline
		Dataset  & \multicolumn{12}{c}{Transportation}                                                                                                                                                          \\ \hline
		\hline
		CIFAR-10 & \multicolumn{3}{c}{Airplane(0\tnote{1})}           & \multicolumn{3}{c}{Automobile(1)}          & \multicolumn{3}{c}{Ship(8)}               & \multicolumn{3}{c}{Truck(9)}                \\
		STL-10 & \multicolumn{3}{c}{Airplane(0)}           & \multicolumn{3}{c}{Car(2)}          & \multicolumn{3}{c}{Ship(8)}               & \multicolumn{3}{c}{Truck(9)}                \\ \hline \hline
		Dataset  & \multicolumn{12}{c}{Animals}                                                                                                                                                          \\ \hline
		\hline
		CIFAR-10 & \multicolumn{2}{c}{Bird(2)} & \multicolumn{2}{c}{Cat(3)} & \multicolumn{2}{c}{Deer(4)} & \multicolumn{2}{c}{Dog(5)} & \multicolumn{2}{c}{Frog(6)} & \multicolumn{2}{c}{Horse(7)} \\
		STL-10 & \multicolumn{2}{c}{Bird(1)} & \multicolumn{2}{c}{Cat(3)} & \multicolumn{2}{c}{Deer(4)} & \multicolumn{2}{c}{Dog(5)} & \multicolumn{2}{c}{Horse(6)} & \multicolumn{2}{c}{Monkey(7)} \\ \hline
	\end{tabular}
	\begin{tablenotes}
		\scriptsize
		\item[1] The number in brackets represents the numeric label of the class in the dataset.
	\end{tablenotes}
\end{table}

\begin{definition}
	(\textbf{Robust Class and Vulnerable Class}) A class whose robustness is higher than the overall robustness is called a robust class. In contrast, a class whose robustness is lower than the overall robustness is called a vulnerable class.
\end{definition}

\section{Class-wise Robustness Analysis} \label{sec4}

In this section, we focus on analyzing the class-wise robustness, including class-biased learning and class-relation exploring on six benchmark datasets. Moreover, we investigate the class-wise robustness with different attack and defense models.

We use six benchmark datasets in adversarial training to obtain the corresponding robust model, \textit{i.e.}, MNIST \citep{lecun1998gradient}, CIFAR-10 \& CIFAR-100 \citep{krizhevsky2009learning}, SVHN \citep{netzer2011reading}, STL-10 \citep{coates2011analysis} and ImageNet \citep{deng2009imagenet}. 
Table \ref{table0} highlights that the classes of CIFAR-10 and STL-10 can be grouped into two superclasses: \textit{Transportation} and \textit{Animals}.
Similarly, CIFAR-100 also contains 20 superclasses with each has 5
subclasses. For the ImageNet dataset, the pipeline of adversarial training follows \citet{wong2019fast}, while the training methods of other datasets follow \citet{madry2018towards}.
See Appendix \ref{A1} for detailed experimental settings.

\subsection{Class-biased Learning}
Figure \ref{fig1} plots the robustness of each class at different epochs in the test set for six benchmark datasets with adversarial training, where the shaded area in each sub-figure represents the robustness gap between different classes across epochs.
Considering the large number of classes in CIFAR-100 and ImageNet, we randomly sample 12 classes for a better indication. From Figure \ref{fig1}, we surprisingly find that there are recognizable robustness gaps between different classes for all datasets.
Specifically, for SVHN, CIFAR-10, STL-10 and CIFAR-100, 
the class-wise robustness gaps are obvious and the largest gaps can reach 40\%-50\% (Figure \ref{fig1-CIFAR-10}-\ref{fig1-STL-10}).
For ImageNet, since the model uses the three-stage training method \citep{wong2019fast}, its class-wise robustness gap increases with the training epoch, and finally up to 80\% (Figure \ref{fig1-ImageNet}). 
Even for the simplest dataset MNIST, on which model has achieved more than 95\% overall robustness, the largest class-wise robustness gap still has 6\% (Figure \ref{fig1-MNIST}).

Hence, we can conclude that the class-bias learning phenomenon is also common in adversarial learning, and there are remarkable robustness discrepancies among classes, leading to unbalance/unfair class-wise robustness in adversarial training. Inspired by the phenomenon in Figure \ref{fig1}, we next conduct the analysis to the robust relation between classes and the impact of different attacks and defenses on class-wise robustness in the following subsections.

\subsection{The relations among different classes} \label{sec4.2}

We first systematically investigate the relation of different classes under robust models.
Figure \ref{fig2} shows the confusion matrices of robustness between classes on all the six datasets. The X-axis and Y-axis represent the predicted classes and the ground truth classes, respectively. The grids on the main diagonal line represent the robustness of each class, while the grids on the off-diagonal line represent the non-robustness on one class (Y-axis) to be misclassified to another class (X-axis). 

\textbf{Observations and Analysis.} From the results reported in Figure \ref{fig2}, we have the following observations and analysis: (i) The confusion matrices on all six benchmark datasets roughly demonstrate one kind of symmetry (\textit{i.e}, the highlight colors of off-diagonal elements are symmetrical about the main diagonal), which indicates that some classes-pair could be easily misclassified between each other.
\begin{figure*}[htbp]
	\centering
	\begin{minipage}{.55\columnwidth}
		\subfloat[Class-wise variance of confidence (CVC) of SOTA defense models]{
			\label{fig3-1-new} 
			\includegraphics[width=.95\columnwidth]{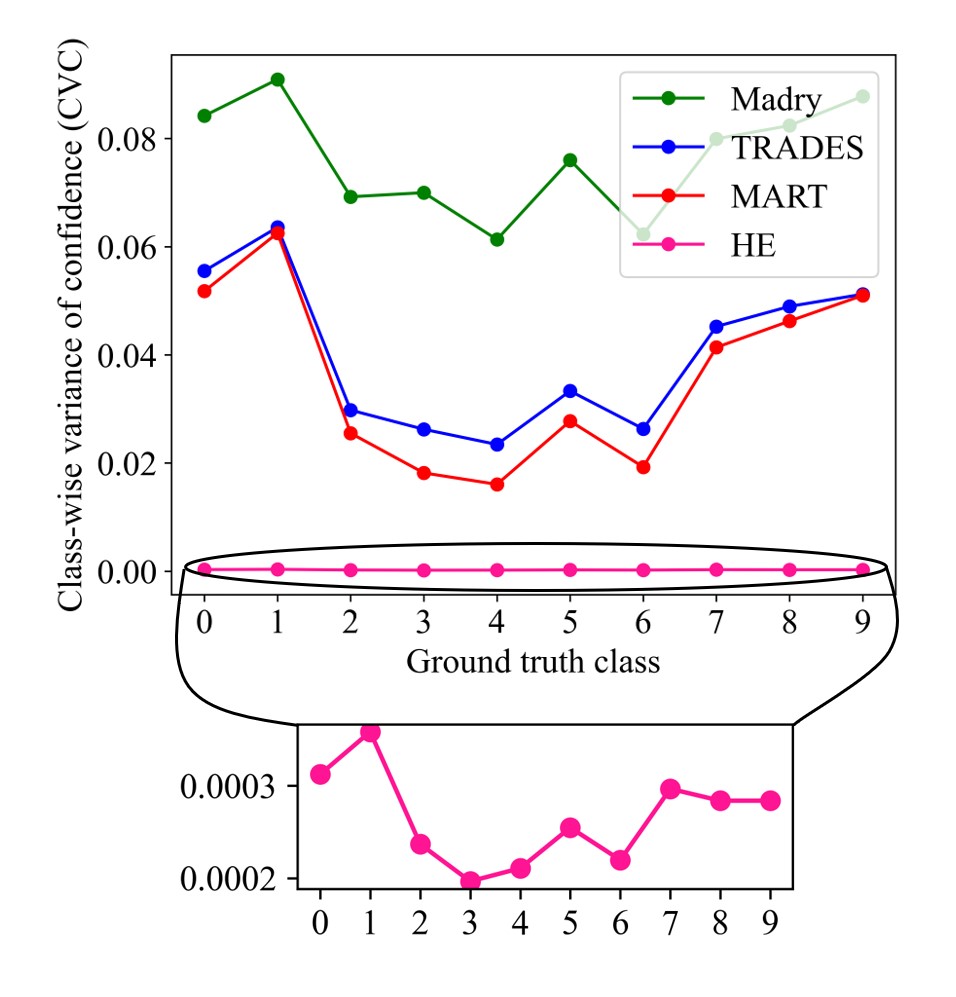}}
	\end{minipage}
	\begin{minipage}{.5\columnwidth}
		\parbox[][4.6cm][c]{\textwidth}{
			\centering
			\includegraphics[width=\textwidth]{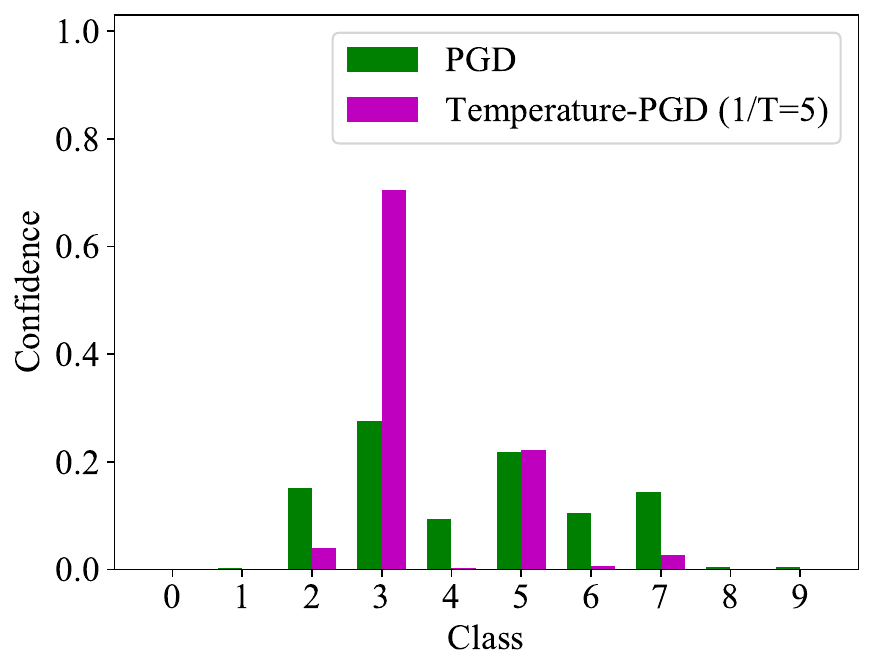}
		}
		\subcaption{MART's output for image 127 (class 3) with iteration steps 1}
		\label{fig3-2-new}
	\end{minipage}
	\begin{minipage}{.5\columnwidth}
		\parbox[][4.6cm][c]{\textwidth}{
			\centering
			\includegraphics[width=\textwidth]{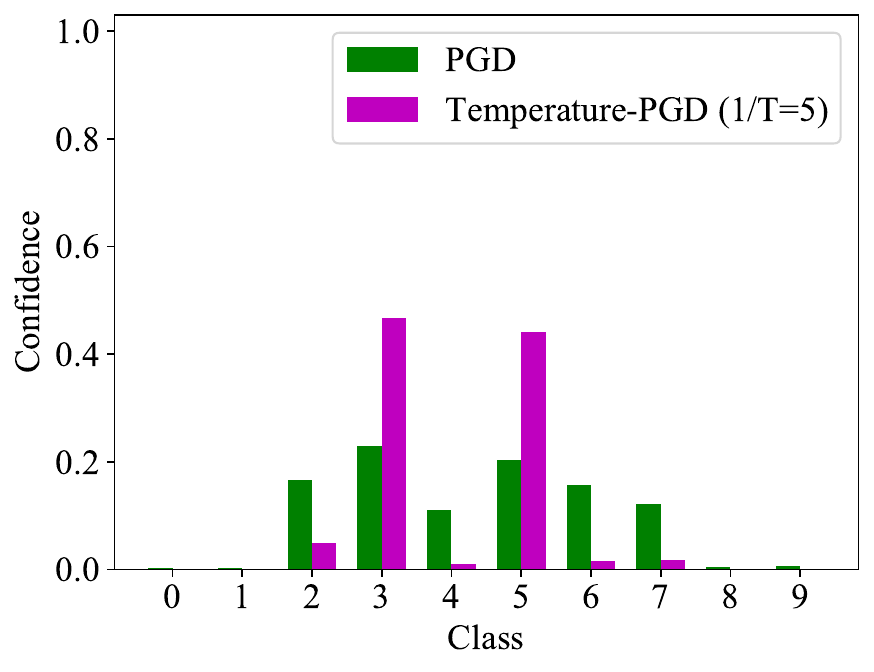}
		}
		\subcaption{MART's output for image 127 (class 3) with iteration steps 10}
		\label{fig3-3-new}
	\end{minipage}
	\begin{minipage}{.5\columnwidth}
		\parbox[][4.6cm][c]{\textwidth}{
			\centering
			\includegraphics[width=\textwidth]{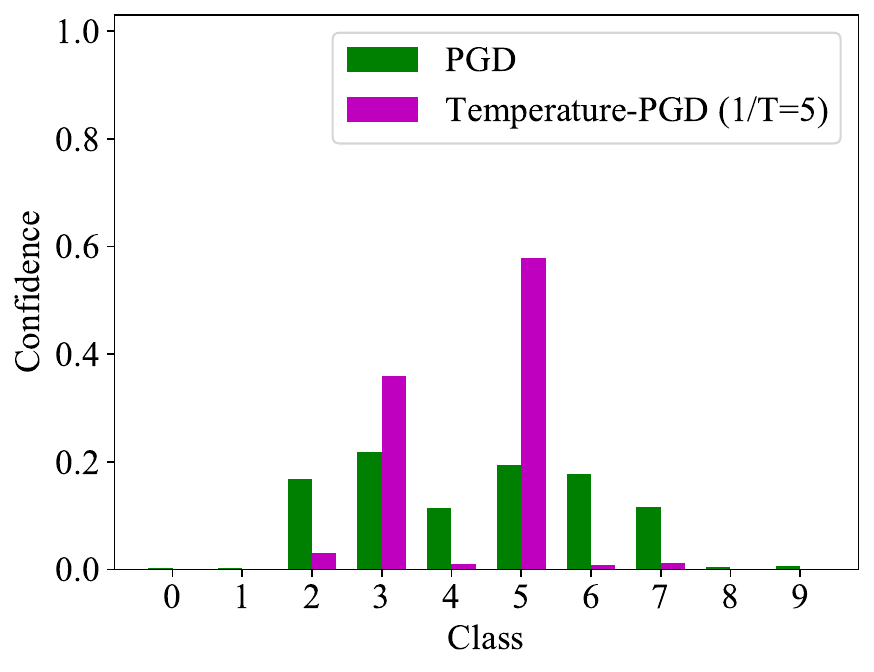}
		}
		\subcaption{MART's output for image 127 (class 3) with iteration steps 20}
		\label{fig3-4-new}
	\end{minipage}
	\caption{Analysis of output confidence}
	\label{fig3-new}
\end{figure*}
\begin{table*}[htbp]
	\tiny
	\centering
	\caption{Adversarial robustness (\%) under Temperature-PGD$^{20}$ attack on CIFAR-10.}
	\label{table4-new}
	\begin{threeparttable}
		\begin{tabular}{ccccccccccccccc}
			\toprule
			Defense&1/T & Tot.  &  0    &  1    &  2    & 3     & 4     & 5     & 6     & 7     & 8     & 9   & CV & MCD \\ \midrule
			Madry&2&    51.8\textcolor{orange}{(-0.3)\tnote{1}}&63.4\textcolor{orange}{(-0.4)}&72.0\textcolor{blue}{(+0.4)\tnote{2}}&38.8\textcolor{orange}{(-0.3)}&25.2\textcolor{orange}{(-0.1)}&33.9\textcolor{orange}{(-2.8)}&38.5\textcolor{orange}{(-0.1)}&56.6\textcolor{orange}{(-0.8)}&59.8\textcolor{blue}{(+0.3)}&63.0\textcolor{orange}{(-0.1)}&66.9\textcolor{blue}{(+0.1)}&235.6\textcolor{blue}{(+11.3)}&46.8\textcolor{blue}{(+0.5)} \\ 
			TRADES&5 & 54.6\textcolor{orange}{(-1.7)}&66.8\textcolor{orange}{(-1.0)}&80.0\textcolor{orange}{(-0.6)}&36.7\textcolor{orange}{(-1.1)}&26.2\textcolor{orange}{(-3.2)}&35.6\textcolor{orange}{(-5.0)}&43.0\textcolor{orange}{(-0.9)}&56.0\textcolor{orange}{(-3.3)}&66.0\textcolor{orange}{(-0.9)}&70.8\textcolor{orange}{(-1.0)}&64.9\textcolor{orange}{(-0.7)}&291.6\textcolor{blue}{(+28.0)}&53.8\textcolor{blue}{(+2.7)}   \\ 
			MART&5&
			54.3\textcolor{orange}{(-3.9)}&62.7\textcolor{orange}{(-1.8)}&77.1\textcolor{orange}{(-0.9)}&41.5\textcolor{orange}{(-3.6)}&26.3\textcolor{orange}{(-9.1)}&27.5\textcolor{orange}{(-10.2)}&41.5\textcolor{orange}{(-2.0)}&60.8\textcolor{orange}{(-4.5)}&66.1\textcolor{orange}{(-1.4)}&72.8\textcolor{orange}{(-3.5)}&67.3\textcolor{orange}{(-2.2)}&311.3\textcolor{blue}{(+76.2)}&50.8\textcolor{blue}{(+8.2)} \\ 
			HE&5& 57.3\textcolor{orange}{(-3.4)}&62.4\textcolor{orange}{(-2.5)}&74.8\textcolor{orange}{(-4.5)}&38.4\textcolor{orange}{(-2.6)}&29.4\textcolor{orange}{(-5.1)}&43.1\textcolor{orange}{(-4.8)}&47.8\textcolor{orange}{(-3.7)}&62.9\textcolor{orange}{(-4.7)}&69.1\textcolor{orange}{(-1.4)}&74.5\textcolor{orange}{(-2.4)}&70.8\textcolor{orange}{(-2.4)}&240.8\textcolor{blue}{(+16.0)}&45.4\textcolor{blue}{(+0.6)} \\
			HE&50&
			50.4\textcolor{orange}{(-10.3)}&58.2\textcolor{orange}{(-6.7)}&71.8\textcolor{orange}{(-7.5)}&33.3\textcolor{orange}{(-7.7)}&17.6\textcolor{orange}{(-16.9)}&23.0\textcolor{orange}{(-24.9)}&41.6\textcolor{orange}{(-9.9)}&56.2\textcolor{orange}{(-11.4)}&66.9\textcolor{orange}{(-3.6)}&69.9\textcolor{orange}{(-7.0)}&65.7\textcolor{orange}{(-7.5)}&363.5\textcolor{blue}{(+138.7)}&54.1\textcolor{blue}{(+9.3)}
			\\ \bottomrule
		\end{tabular}
		\begin{tablenotes}
			\tiny
			\item[1] “-” represents the robustness reduction compared with the corresponding element of PGD attack in table \ref{table4}.
			\item[2] “+” represents the robustness improvement compared with the corresponding element of PGD attack in table \ref{table4}.
		\end{tablenotes}
	\end{threeparttable}
\end{table*}
(ii) The symmetry classes-pair in Figure \ref{fig2} are always similar to some degree, such as similar in shape or belonging to the same superclasses, hence, would be easy misclassified to each other.
Specifically, for SVHN, digits with similar shapes are more likely to be flipped to each other, \textit{e.g.}, the number 6 and number 8 are similar in shape and the non-robustness between them (number 6 is misclassified to be number 8 or vice versa) is very high as shown in Figure \ref{fig2-SVHN}.
For CIFAR-10 and STL-10, Figures \ref{fig2-CIFAR-10} and \ref{fig2-STL-10} clearly show that the classes belonging to the same superclass have high probabilities to be misclassified to each other, for example, both class 3 (cat) and class 5 (dog) in CIFAR-10 belong to the superclass \textit{Animals}, the non-robustness between them is very high in Figure \ref{fig2-CIFAR-10}. (iii) Few misclassifications would happen between two classes with different superclasses.
For example, in STL-10, the class 5 (dog) belongs to superclass \textit{Animals}, while class 9 (truck) belongs to \textit{Transportation}, and their non-robustness is almost 0 as shown in figure \ref{fig2-STL-10}.

For CIFAR-100 and ImageNet, we can also observe symmetry properties of confusion matrix in Figure \ref{fig2-CIFAR-100} and Figure \ref{fig2-ImageNet}, which is consistent with the above analysis.
Overall, Figure \ref{fig2} demonstrates that the classes with similar semantic would be easier misclassified (with higher non-robustness) to each other than those with different semantics (\textit{e.g.}, the classes belong to different superclasses).

\subsection{The class-wise robustness under different attacks and defenses} \label{sec4.4}

The above analysis mainly concentrates on the performance under ${\rm PGD}$ attack. In this subsection, we investigate the class-wise robustness of state-of-the-art robust models against various popular attacks in the CIFAR-10 dataset. 

The defense methods we chose include Madry training \citep{madry2018towards}, TRADES \citep{zhang2019theoretically}, MART \citep{wang2019improving} and HE \citep{pang2020boosting}. 
We train WideResNet-32-10 \citep{zagoruyko2016wide} following
the original papers. 
White-box attacks include FGSM \citep{goodfellow2014explaining}, ${\rm PGD}$ \citep{madry2018towards} and $\rm CW_{\infty}$ \citep{carlini2017towards}, and the implementation of $\rm CW_{\infty}$ follows \citep{carmon2019unlabeled}.
Black-box attacks include a transfer-based and a query-based attack. 
The former uses a standard trained WideResNet-32-10 as the substitute model to craft adversarial examples, and the latter uses $\mathcal{N}$ attack \citep{li2019nattack}.
All hyperparameters see Appendix \ref{A2}.

In order to quantitatively measure the robustness unbalance (or discrepancy) among classes, we give the definition of two statistical metrics: class-wise variance (CV) and maximum class-wise discrepancy (MCD) as follows
\begin{figure*}[htbp]
	\centering
	
	\begin{subfigure}[b]{0.32\textwidth}
		\centering
		\includegraphics[width=\textwidth]{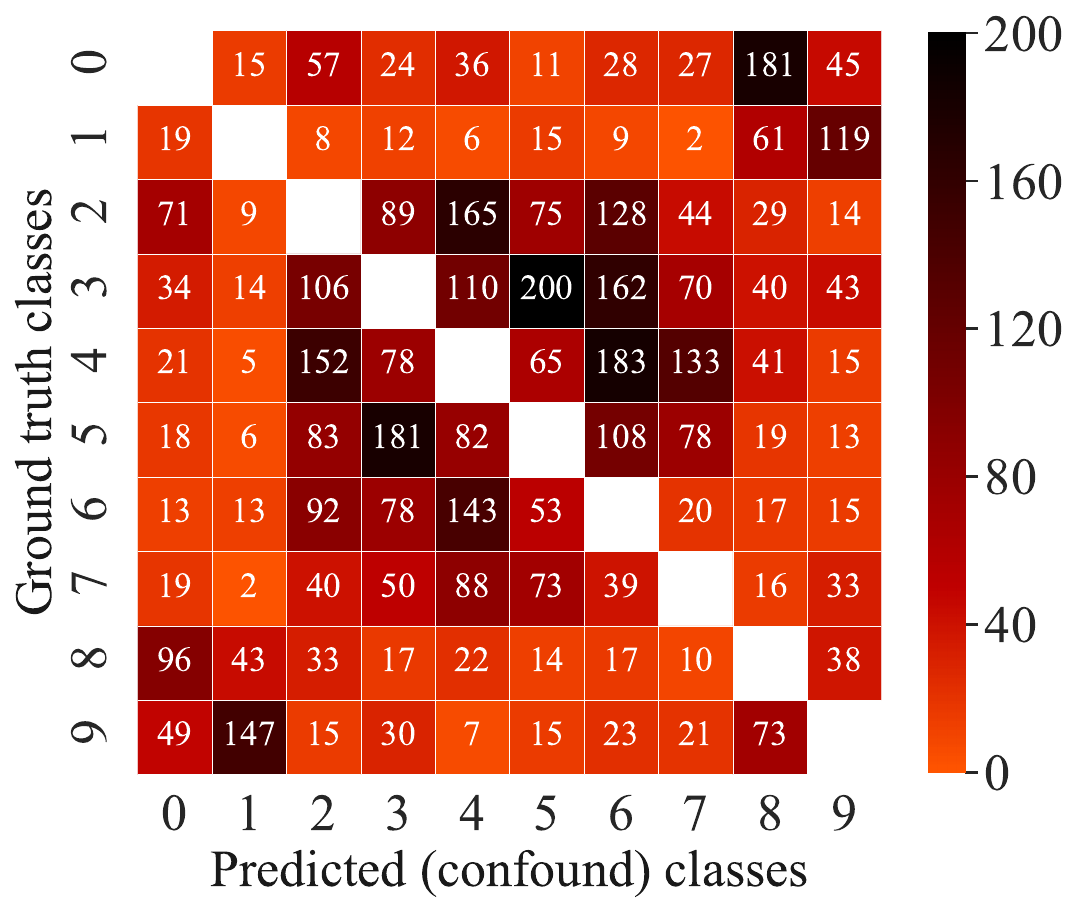}
		\caption{Misclassified confusion matrix}
		\label{fig3-Orig}
	\end{subfigure} 
	\hspace{0.01\textwidth}
	\begin{subfigure}[b]{0.32\textwidth}
		\centering
		\includegraphics[width=\textwidth]{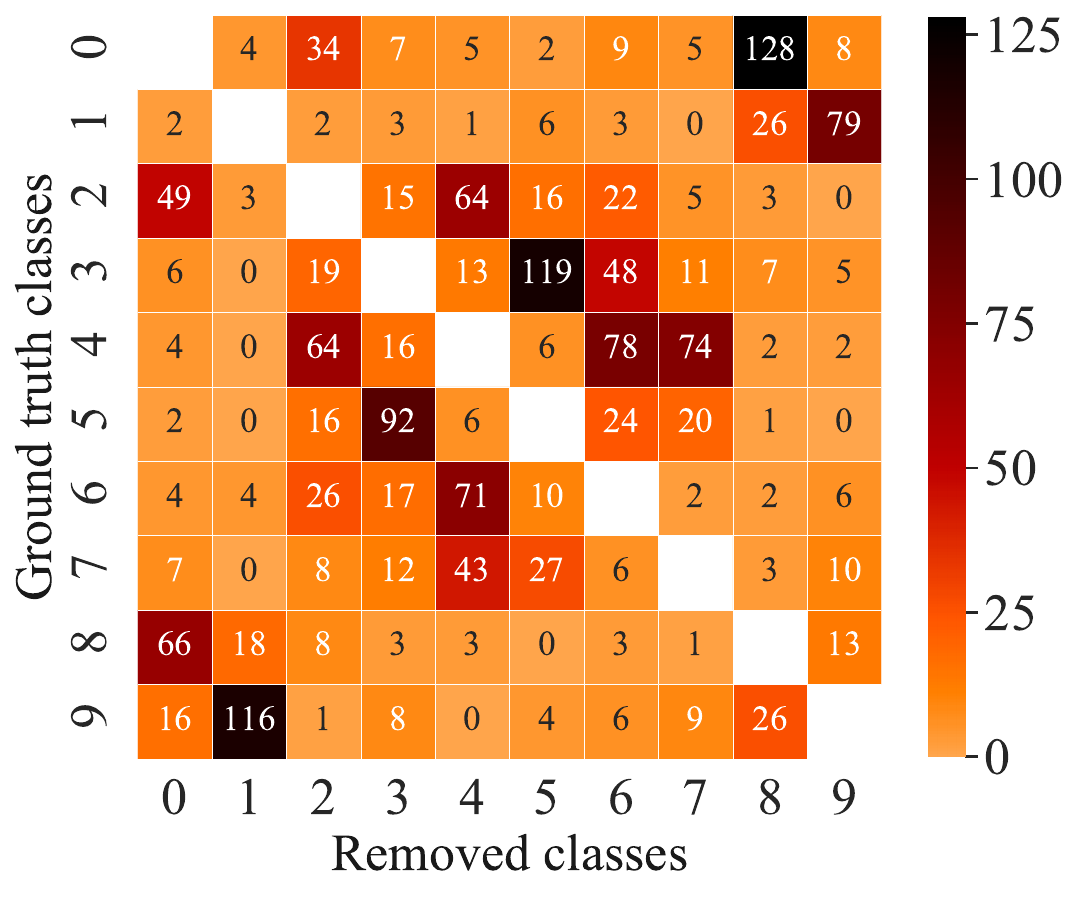}
		\caption{Homing confusion matrix}
		\label{fig3-Drop}
	\end{subfigure}
	\hspace{0.01\textwidth}
	\begin{subfigure}[b]{0.32\textwidth}
		\parbox[][4.5cm][c]{\textwidth}{
			\centering
			\includegraphics[width=\textwidth]{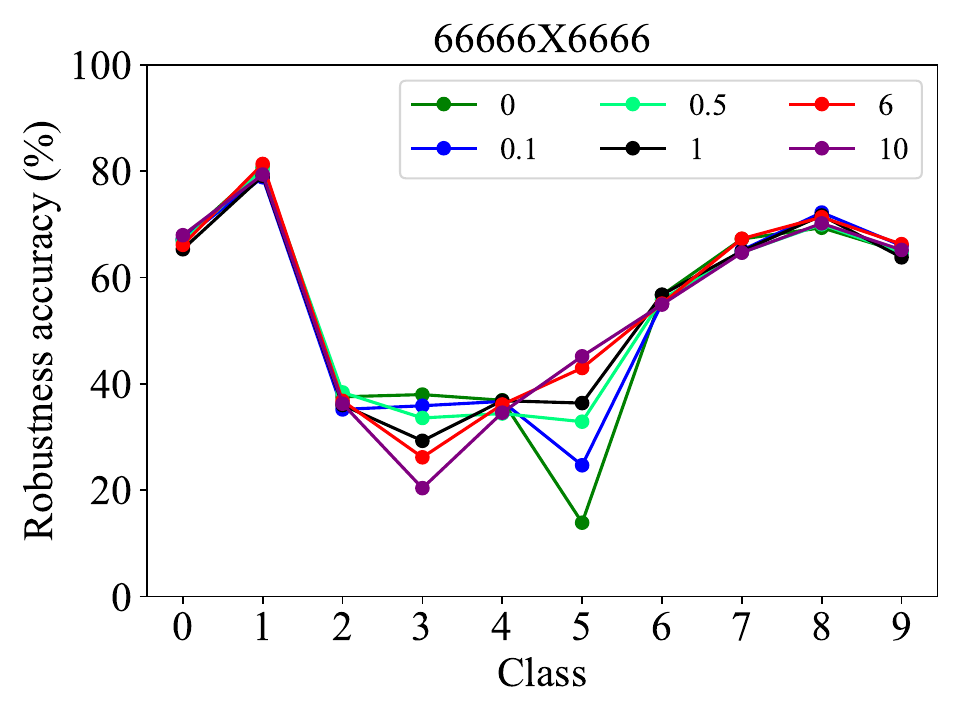}
		}
		\subcaption{Adjust class 3 robustness}
		\label{fig3-Adj1}
	\end{subfigure}
	\caption{Case study about adjusting class 3 robustness in the training phase}
	\label{fig3}
\end{figure*}

\begin{definition}
	(\textbf{Class-wise Variance ,CV}) Given one dataset containing $C$ classes, 
	the accuracy of each class $c$ is $a_c$, the average accuracy over all classes is $\overline{a}= \sum_{c=1}^C a_c/C$, and then CV is defined as:
	$$CV = \frac{1}{C} \sum_{c=1}^C(a_c-\overline{a})^2.$$
\end{definition}
\begin{definition}
	(\textbf{Maximum Class-wise Discrepancy, MCD}) Given one dataset, let $a_{max}$ and $a_{min}$ represent the maximum and minimum accuracy of class, then MCD is defined as:
	$$MCD = a_{max}-a_{min}.$$
\end{definition}
The insight of these two metrics is to measure the average discrepancy and the most extreme discrepancy among classes.
Intuitively, these metrics will be large if there are huge class-wise differences. 

\textbf{Observations and Analysis.} Based on the CIFAR-10 dataset, we check the class-wise robustness of different attack and defense models and report the results in Table \ref{table4}. From the results, we can have the following observations and analysis: (i) In all models and attacks, there are remarkable robustness gaps between different classes, and class 1 and class 3 are always the most robust and vulnerable class in all settings, which suggests the relative robustness of each class has a strong correlation with the dataset itself. (ii) Stronger attacks in white-box settings are usually more effective for vulnerable classes. For example, comparing FGSM and PGD of the same defense method, the robustness reduction of the vulnerable classes (\textit{e.g.}, class 3) is obviously larger than that of robust classes (\textit{e.g.}, class 1), resulting in larger class-wise variance (CV) and maximum class-wise discrepancy (MCD). (iii) In black-box settings, the main advantage of the query-based attack over the transfer-based attack is also concentrated in vulnerable classes.
One explanation is that many examples of these classes are closer to the decision boundary, making it easier to be attacked.

In addition, we have also checked the CV and MCD of the adversarial training are significantly larger than the standard training in all datasets. For example, in terms of the CIFAR-10 dataset, the CV of adversarial training is 28 times that of standard training, and the MCD of adversarial training is 5 times that of standard training. This shows that class-wise properties in the robustness model are worthy of attention. See Appendix \ref{A3} for more details.
%

\section{Improving adversarial attack via class-wise discrepancies}
Although Section \ref{sec4.4} have shown the class-wise robustness discrepancies are commonly observed in adversarial settings, we believe that this gap can be further enlarged if the attacker makes full use of the properties of vulnerable classes. Specifically,
since the images near decision boundary usually have smooth confidence distributions, popular attacks cannot find the effective direction in the iterative process, and
Figure \ref{fig3-2-new}-\ref{fig3-4-new} clearly show an example of the failed attack with PGD (\textit{i.e.}, the bar for ground truth class 3 is always the highest).
To solve this problem, we propose to use a temperature factor to change this distribution, so as to create \textit{virtual power} in the possible adversarial direction.

For a better formulation, we assume that the DNN is $f$, the input example is $x$, the number of classes in the dataset is $C$, then the softmax probability of this sample $x$ corresponding to class $k$ ($k \in C$) is
\begin{align}
	\mathbb{S}(f(x))_{k}=\frac{e^{f(x)_{k}/T}}{\sum_{c=1}^{C}e^{f(x)_{c}/T}}. 
\end{align}
Using this improved softmax function, the adversarial perturbation crafted at $t^{\rm th}$ step is
\begin{align}
	\textstyle
	\delta^{t+1}=\prod_{\epsilon}(\delta^{t}+\alpha \cdot {\rm sign}(\nabla \ell(\mathbb{S}(f(x+\delta^{t})),y))).
\end{align}
\begin{figure*}[htbp]
	
	\centering
	\begin{subfigure}[b]{0.32\textwidth}
		\parbox[][4.5cm][c]{\textwidth}{
			\centering
			\includegraphics[width=\textwidth]{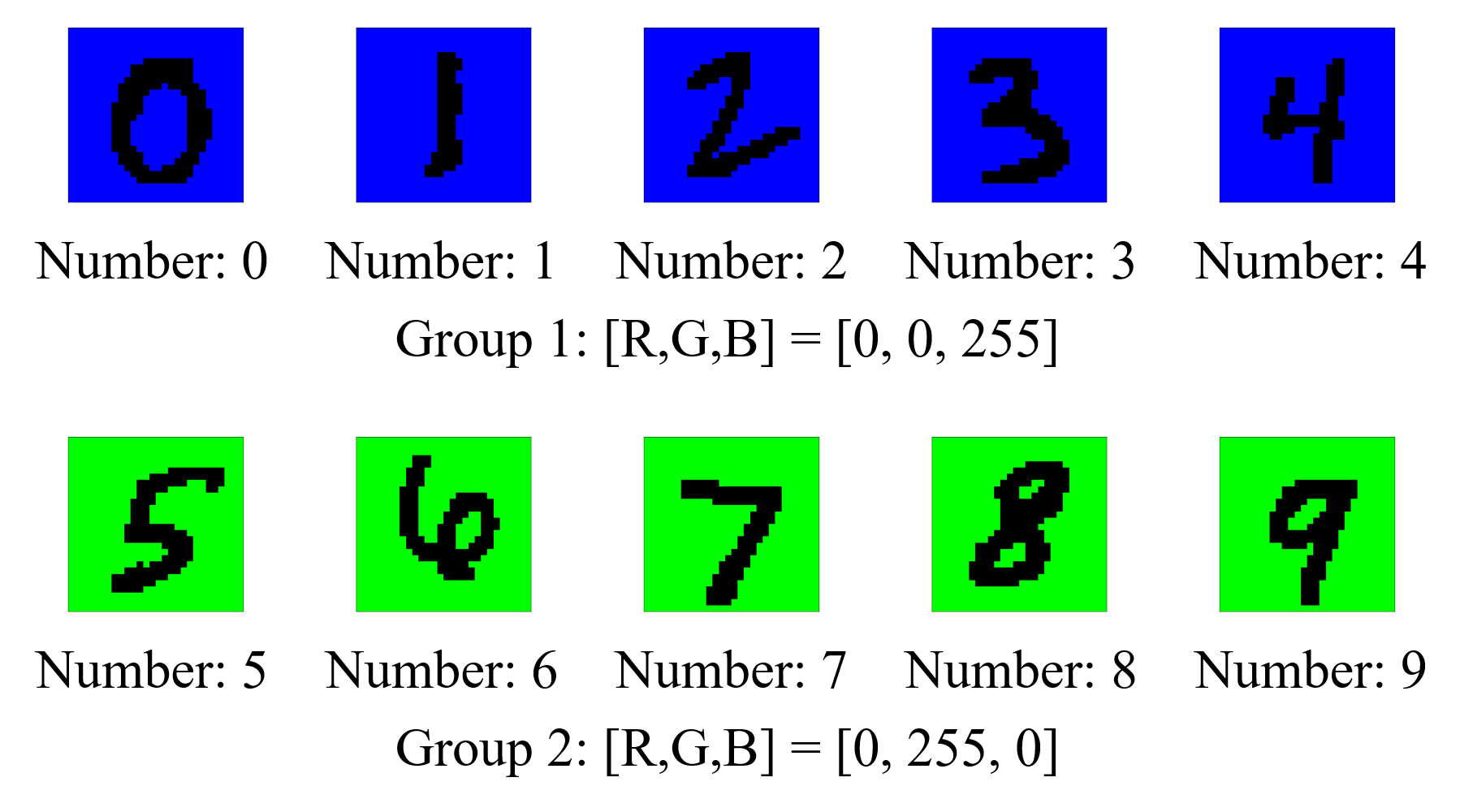}
		}
		\subcaption{Numbers after adding background}
		\label{fig4-1}
	\end{subfigure} \hfill
	\begin{subfigure}[b]{0.32\textwidth}
		\centering
		\includegraphics[width=\textwidth]{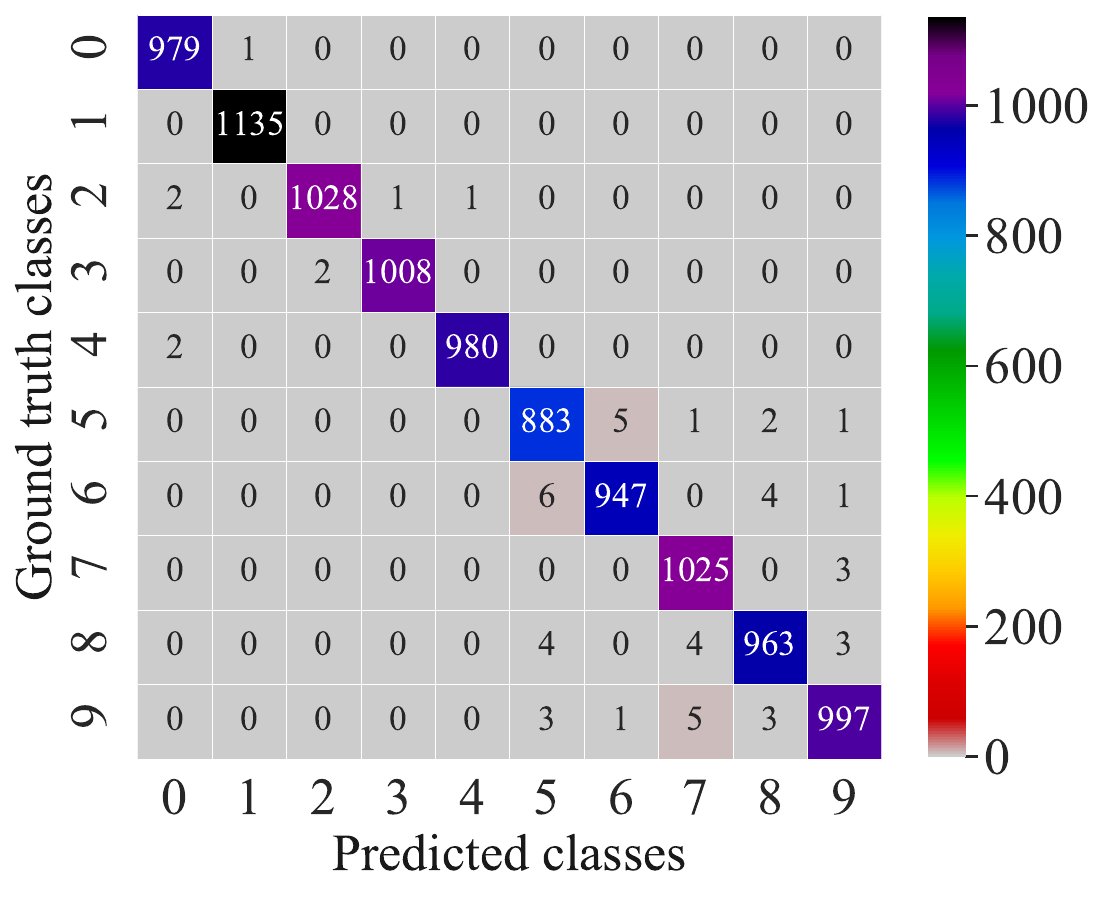}
		\caption{Confusion matrix for natural images}
		
		\label{fig4-2}
	\end{subfigure} \hfill
	\begin{subfigure}[b]{0.32\textwidth}
		\centering
		\includegraphics[width=\textwidth]{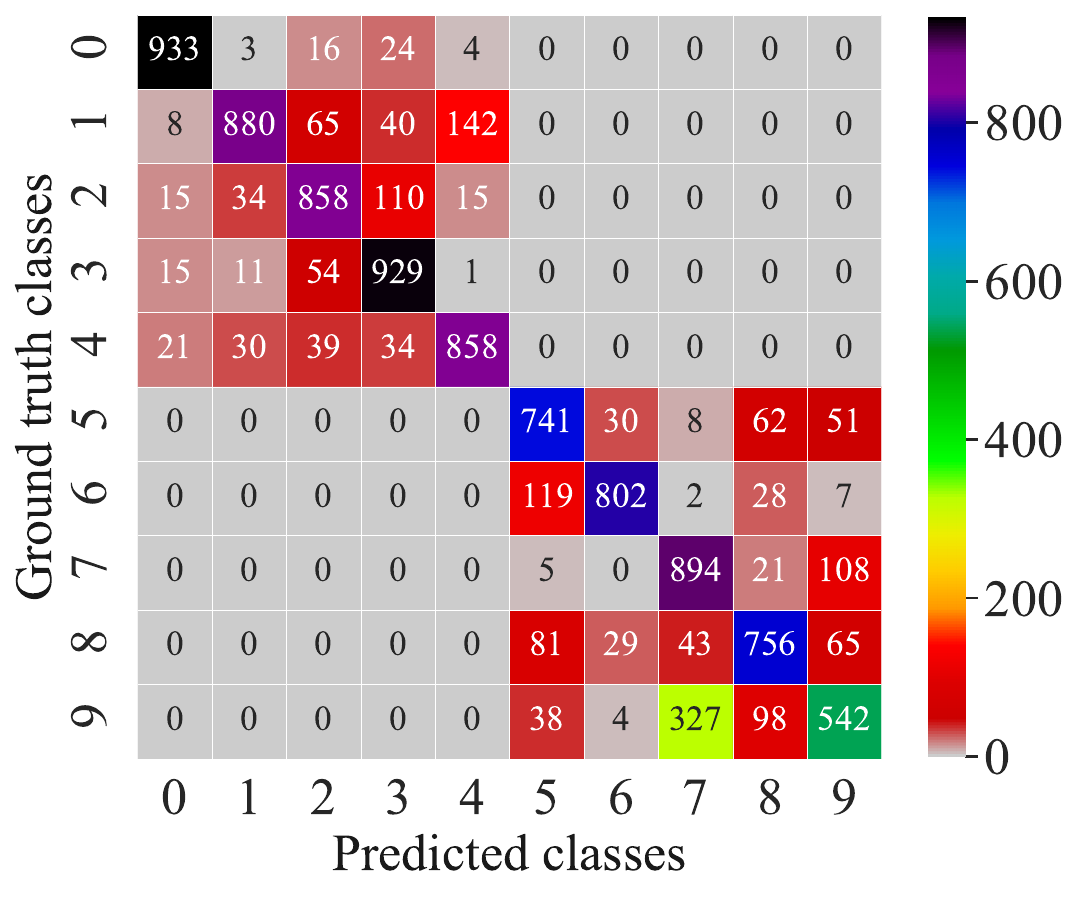}
		\caption{Confusion matrix for adversarial images}
		
		\label{fig4-3}
	\end{subfigure}
	\caption{Experiments about adding background on MNIST}
	\label{fig4}
\end{figure*}
Where $\prod_{\epsilon}$ is the projection operation, which ensures that $\delta$ is in $\epsilon$-ball. $\ell(\cdot)$ is the cross-entropy loss. $\alpha$ is the step size. $y$ is the ground truth class.

The bar corresponding to Temperature-PGD (1/T=5) in Figure \ref{fig3-2-new}-\ref{fig3-4-new} is a good example of how our proposed method works. 
To better understand the impact of our method on different defense models, 
the class-wise variance of confidence (CVC) is proposed to measure the smoothness of the confidence output of these models.

\begin{definition}
	(\textbf{Class-wise Variance of Confidence, CVC}) Assume that there are $C$ classes in the test set, class $k$ has $N$ images, the confidence output of one image $i$ is $\bm{p}=(p_1,\cdots,p_c,\cdots,p_C)$ and the average confidence of this image is $\overline{p}^i=\sum_{c=1}^C p_c^i /C$, then CVC of class $k$ is defined as $$CVC_k=\frac{1}{NC}\sum _{i=1}^N\sum _{c=1}^C (p_c^i-\overline{p}^i)^2.$$
\end{definition}
\begin{figure}[t]
	\centering
	\begin{subfigure}[b]{0.44\textwidth}
		
		\centering
		\includegraphics[width=\textwidth]{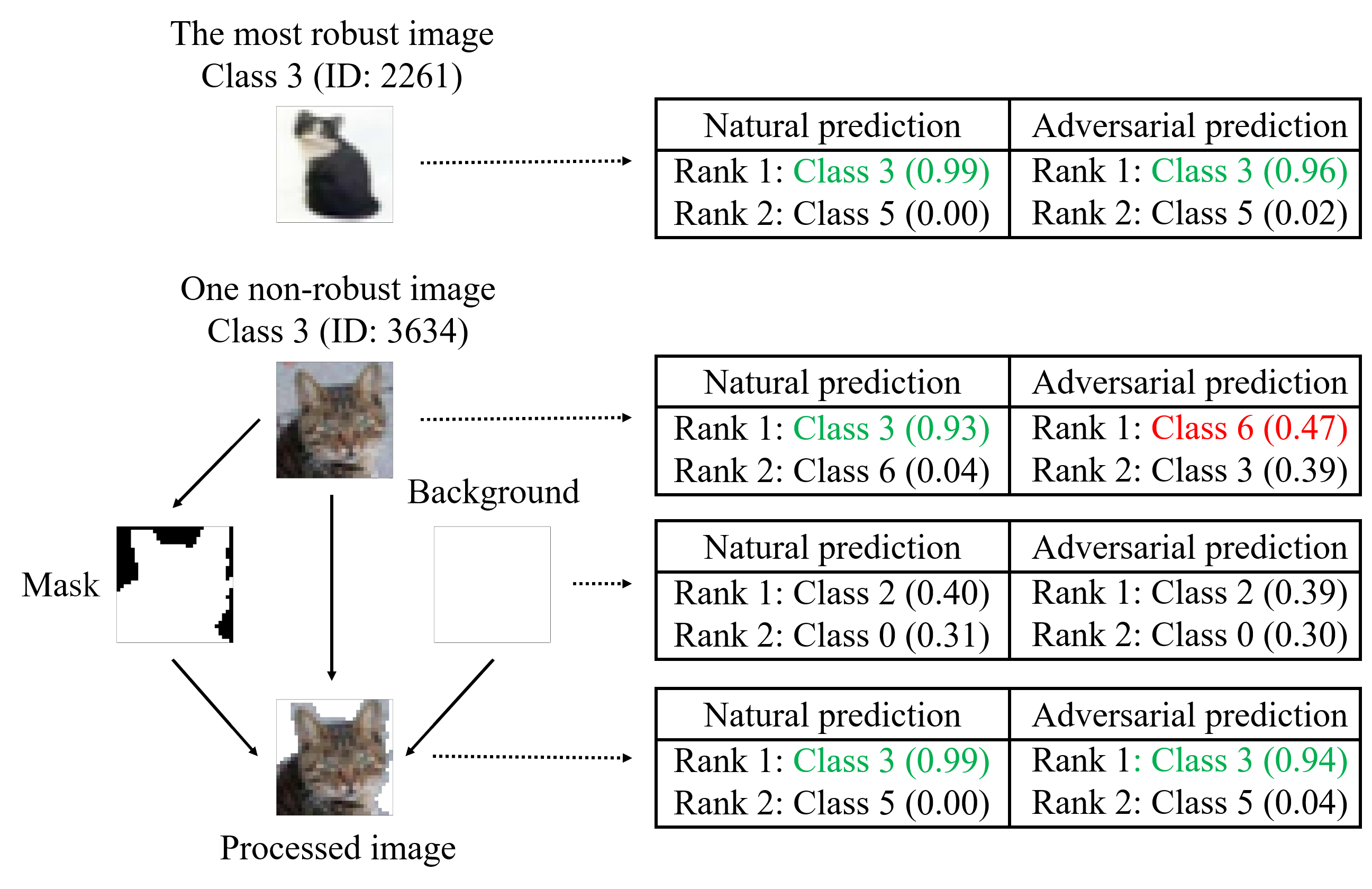}
	\end{subfigure} \hfill
	\caption{Changing background of class 3 images in CIFAR-10}
	\label{fig5}
\end{figure}

Intuitively, this value will be small if the output confidence of one class is smooth. 
From the results of Figure \ref{fig3-1-new} and Table \ref{table0}, we can find that in all defense models, the CVC of superclass \textit{Animals} is smaller than that of superclass \textit{Transportation}, and the CVC of class 4 and class 3 is the smallest and second-smallest. Combined with the information of Table \ref{table4}, the class with low robustness is closer to the classification boundary, so the confidence distribution is smoother, which is consistent with our previous analysis. On the other hand, the overall CVC of HE is much smaller than other defense methods, which means that popular attacks (\textit{i.e.,} PGD) may be very inefficient for this defense model.

\textbf{Overall results.}
In practice, we perform a grid search on the hyperparameter $\rm{1/T}\in[2,5,10,50]$, and report the best performance.
The results of Table \ref{table4-new} verify the effectiveness of our proposed method. 
Specifically, (i) The CV and MCD of all defense models have become larger, which means that the disparity between the classes is enlarged. (ii) Since the confidence output of the vulnerable classes is smoother (Figure \ref{fig3-1-new}), our method can significantly reduce the robustness of these classes. (\textit{e.g.}, class 3 and class 4). (iii) The Madry’s model has a steeper confidence distribution, so it is not sensitive to Temperature-PGD attack. On the contrary, because the confidence outcomes of the HE’s model is extremely smooth, increasing the temperature factor to make the outcome steeper can significantly improve the attack rate, \textit{i.e.}, total robustness reduces 10.3\% (when 1/T=50), of which class 3 and class 4 reduce 16.9\% and 24.9\% respectively.
Overall, the success of Temperature-PGD is effective evidence that stronger attackers can further increase the class-wise robustness difference.

\section{Improving the robustness of the vulnerable class}

In this section, we propose two methods to mitigate the difference in class-wise robustness.
Specifically, our goal is to improve the robustness of the most vulnerable subgroup in CIFAR-10 (\textit{i.e.}, class 3), because it has the lowest robustness as described in the previous analysis.

\subsection{Adjust robustness at the training phase} \label{sec5.1}
Figure \ref{fig2} analyzes the relation of class-wise robustness in detail. Here we further explore the more fine-grained relation between these classes by removing the confound class (Definition \ref{def_2}).
Specifically, for the example $\bm{x}$ from class $y$ is attacked to the confound class $y^\prime$, we are curious if we remove confound class $y^\prime$ (\textit{i.e.,} remove all examples of ground truth class $y^\prime$ in the training set) and re-train the model, will example $\bm{x}$ become a robust example WITHOUT being maliciously flipped to a new confound class\footnote{We usually think that there are many decision boundaries in bounded $\epsilon$-ball, so a new confound class is likely to appear even if one decision boundary is removed.}?

\begin{definition}
	(\textbf{Homing Property}) Given an adversarial example $x'$ from class $y$ which is misclassified as the confound class $y^\prime$ by a model, this example satisfies homing property if it becomes a robust example after we re-train the model via removing confound class $y^\prime$.
\end{definition}
\begin{table*}[t]
	\scriptsize
	\centering
	\caption{Robust model prediction results for images of class 3 in the test set
		(1000 images) under Temperature-PGD$^{20}$ attack.}
	\label{table3}
	\begin{threeparttable}
		\begin{tabular}{cccccccccccc}
			\toprule
			Line number & Test set                       & Class 0              & Class 1              & Class 2              & \textbf{Class 3(correct)}     & Class 4              & Class 5              & Class 6              & Class 7              & Class 8              & Class 9       \\      
			\midrule  
			1& Original image (Natural)         & 18\tnote{1}                   & 4                    & 43                   & \textbf{710}                  & 38                   & 88                  & 64                  & 15                   & 6                   & 14    \\               
			
			2& + white background (Natural)     & 39\textcolor{blue}{(+21)\tnote{2}}              & 10\textcolor{blue}{(+6)}                & 48\textcolor{blue}{(+5)}              & \textbf{742\textcolor{blue}{(+32)}}             & 13\textcolor{orange}{(-25)}              & 59\textcolor{orange}{(-29)}              & 63\textcolor{orange}{(-1)}             & 5\textcolor{orange}{(-10)}                & 3\textcolor{orange}{(-3)}               & 
			18\textcolor{blue}{(+4)}   \\    
			3& + training adjustment method (Natural)     & 35\textcolor{blue}{(+17)}              & 15\textcolor{blue}{(+11)}                & 56\textcolor{blue}{(+13)}              & \textbf{758\textcolor{blue}{(+48)}}             & 23\textcolor{orange}{(-15)}              & 24\textcolor{orange}{(-64)}              & 61\textcolor{orange}{(-3)}             & 6\textcolor{orange}{(-9)}                & 8\textcolor{blue}{(+2)}               & 
			14\textcolor{blue}{(+0)}   \\  
			\midrule          
			4&Original image (Adversarial)     & 34                   & 16                   & 82                  & \textbf{262}                  & 113                  & 254                  & 139                  & 53                   & 14                   & 33             \\      
			5&+ white background (Adversarial) & 76\textcolor{blue}{(+42)}              & 22\textcolor{blue}{(+6)}               & 72\textcolor{orange}{(-10)}              & \textbf{403\textcolor{blue}{(+141)}}            & 80\textcolor{orange}{(-33)}              & 136\textcolor{orange}{(-118)}             & 116\textcolor{orange}{(-23)}             & 38\textcolor{orange}{(-15)}              & 7\textcolor{orange}{(-7)}               & 
			50\textcolor{blue}{(+17)}\\
			6&+ training adjustment method (Adversarial) & 93\textcolor{blue}{(+59)}              & 31\textcolor{blue}{(+15)}               & 80\textcolor{orange}{(-2)}              & \textbf{435\textcolor{blue}{(+173)}}            & 72\textcolor{orange}{(-41)}              & 84\textcolor{orange}{(-170)}             & 111\textcolor{orange}{(-28)}             & 35\textcolor{orange}{(-18)}              & 3\textcolor{orange}{(-11)}               & 
			56\textcolor{blue}{(+23)}\\
			\bottomrule
		\end{tabular}
		\begin{tablenotes}
			\scriptsize
			\item[1] The number represents how many images in the corresponding test set are predicted to be the corresponding class.
			\item[2] “+” represents the increase in the number of images compared with the corresponding element of the original image test set, and “-” is vice versa.
		\end{tablenotes}
	\end{threeparttable}
\end{table*}

To explore the above question, we conduct extensive experiments and 
the results are reported in Figure \ref{fig3}.
Figure \ref{fig3-Orig} and Figure \ref{fig2-CIFAR-10} are similar, and the difference is that the values in Figure \ref{fig3-Orig} represent the number of examples instead of percentage, 
and the main diagonal elements (the number of examples correctly classified) are hidden for better visualization and comparison. Thus this figure is called the Misclassified confusion matrix. 
To check the \textit{homing property}, we alternatively remove each confound class to re-train the model and plot the results in Figure \ref{fig3-Drop}, where the element in the $i^{\rm th}$ row and $j^{\rm th}$ column (indexed by the classes starting from $0$) indicates how many adversarial examples with ground truth class $i$ and confound class $j$ that satisfy \textit{homing property} (\textit{i.e.,} these examples will become robust examples after removing the confound class $j$),
so this figure is defined as the Homing confusion matrix.

Figure \ref{fig3} clearly shows \textit{homing property} is widely observed in many misclassified examples.
For example, we can focus on the 3$^{\rm rd}$ row and the 5$^{\rm th}$ column of Figure \ref{fig3-Orig} and \ref{fig3-Drop}. 
200 in Figure \ref{fig3-Orig} means that 200 examples of class 3 are misclassified as class 5, and 119 in Figure \ref{fig3-Drop} means that if we remove class 5 and re-train the model, 119 of 200 examples will \textit{home} to the correct class 3 (\textit{i.e.,} become robust examples). 
This suggests that changing the robustness of class 3 only needs to carefully handle the relation with class 5.
Interestingly, these group-based relations are commonly observed in CIFAR-10, \textit{e.g.}, class 1 (automobile)-class 9 (truck) and class 0 (airplane)-class 8 (ship).

\textbf{The proposed method.}
Based on the above discovery, we try to use this group-based relation to adjust the class-wise robustness. Our method is based on TRADES \citep{zhang2019theoretically} as shown in the Equation (\ref{eq8}).
Specifically, \citet{zhang2019theoretically} set $\beta$ as a constant to adjust  natural accuracy and robust accuracy, while we modify $\beta$ to a vector $\bm{\beta}=(\beta_1,...,\beta_c,...,\beta_C)$ to adjust class-wise robustness, where $c\in C$ is the class id, thus the loss function of class $c$ is

\begin{align}
	\label{eq8}
	\textstyle
	\min\limits_{\bm{\theta}} \frac{1}{n} \sum_{i=1}^{n} \ell\left(h_{\bm{\theta}}\left({\bm{x}}_{c}^i\right), y_{c}^i\right) + \beta_c \max\limits _{\left\|{\bm{x}}_{c}^{\prime,i}-{\bm{x}}_{c}^i\right\|_{p} \leq \epsilon} \mathcal{K} \left(h_{\bm{\theta}}\left({\bm{x}_c^i}\right), h_{\bm{\theta}}\left({\bm{x}}_{c}^{\prime,i}\right)\right). 
\end{align}

Since class 3 and class 5 have an obvious one-to-one relation in Figure \ref{fig3-Drop}.
We only change the $\beta_c$ of class 5 to adjust the robustness of class 3, while fixes the $\beta_c$ of other classes. The result is shown in Figure \ref{fig3-Adj1}. Each line represents the class-wise robustness under the Temperature-PGD attack. The title in the figures represents the $\beta_c$ value of each class $c$, that is, '66666X6666' stands for 
$\beta_c=6$ ($\forall c \in C$ and $c \neq 5$), and the number 6 is chosen to be comparable to the experiment of \citet{zhang2019theoretically} ($\forall c \in C, \beta_c=6$).

\textbf{Overall results.} 
Figure \ref{fig3-Adj1} demonstrates that the robustness of class 3 can be improved or reduced by adjusting the value of $\beta_5$. 
Specifically, when $\beta_5=6 \rightarrow \beta_5=0.5$,
the robustness of class 3 changes from 26.2\% to 33.6\% and the robustness of class 5 changes from 43.0\% to 34.1\%, which shows our method can effectively adjust the robustness of the most vulnerable class 3, thereby reducing the class-wise disparity. Intuitively, Other group-based relations can also be used to further balance the overall robustness.

\subsection{Adjust robustness at the inference phase} 
MNIST and CIFAR-10 are the most commonly used datasets for adversarial training. However, the overall performance of robust models in MNIST usually exceeds 95\%, while this is only 50\%-60\% in CIFAR-10. We speculate that the unified background of MNIST is one of the potential reasons why its performance is better.

To verify our assumption, we add different backgrounds to each class of images in MNIST to explore the role of the background. 
Specifically, we first modify the original images into three-channel images and then add two sets of background colors to the training set and test set of each class, as shown in Figure \ref{fig4-1}. 
In the training and inference phase, $\epsilon$ is set to 0.5 to highlight the robust relation between classes, and other settings are consistent with Section \ref{sec4}.
In addition, we have also verified that this background-changing dataset has almost no effect on the accuracy of standard training. Therefore, we only report the confusion matrices of adversarial training as shown in Figure \ref{fig4-2} and Figure \ref{fig4-3}.

The confusion pattern in Figure \ref{fig4-3} is completely consistent with the background relation of each class in Figure \ref{fig4-1}, which is the evidence that the class-wise robust relation can be changed through the background.
One possible explanation is that the model mistakenly learned the spurious correlation \citep{shen2018causally} between the foreground and the background during the training process, \textit{e.g.}, the model may think that the number 2 and the number 3 are more similar since they have the same background, while the number 2 and the number 5 are vice versa.
However, from the perspective of causality \citep{shen2018causally}, the intrinsic feature to judge whether numbers are similar should be the shape rather than the background.
In fact, \citet{shen2018causally} has proved that this phenomenon has a negative impact on model prediction, but comparing the results of Figure \ref{fig4-2} and Figure \ref{fig4-3}, it is clearly demonstrated that the influence of the background on the adversarial examples is much greater than that on the natural examples, which makes this factor very important in adversarial settings.
To the best of our knowledge, this is the first step to explore the connection between background and model robustness.

\textbf{The proposed method.} 
Inspired by the above phenomenon, 
we believe that the complex background in the CIFAR-10 dataset may affect the robustness of each class and we can use this property to adjust class-wise robustness. To check this, we first select the images with the ground truth class 3 in the test set and then
record the confidence of adversarial prediction corresponding to the class 3 of each image (\textit{i.e.,} $\mathbb{S}(h_{\bm{\theta}}({\bm{x}}_{i}^\prime))_{y=3}$, where $\mathbb{S}$ is the softmax function) and visualize images according to the confidence from high to low.
Surprisingly, we find that the backgrounds of the highly robust images in class 3 are pure white color. Figure \ref{fig5} shows the most robust image (ID: 2261) in class 3 has this white background.

Therefore, we manually extract the mask that can locate the background from one non-robust image (ID: 3634) of class 3 in the test set, and then replace the original background with a white background to investigate the change of prediction.
As shown in Figure \ref{fig5}, the boxes represent the natural and robust prediction of the corresponding image. `Rank 1' and `Rank 2' represent the classes with the highest and the second-highest confidence, and the value in brackets represents the specific confidence.
The result indicates that this non-robust image can become a robust one by replacing the background, while it slightly affects the natural prediction.

We apply the above image processing method to all images of class 3 (1000 images) in the test set to verify whether the above phenomenon can be generalized. 
As shown in Table \ref{table3}.
The number in each row represents how many images in the corresponding test set are predicted to be the corresponding class. 
Since the ground truth of all test images is class 3, the column corresponding to class 3 is the number of images that are correctly predicted.

\textbf{Overall results.} 
As illustrated in Line 4 and Line 5 of Table \ref{table3}, many non-robust examples become robust after adding a white background (\textit{i.e.}, the robustness changed from 26.2\% to 40.3\%), while Line 1 and Line 2 indicate natural predictions are not sensitive to the background, 
which proves that the background mainly has a great influence on the model’s adversarial prediction.
Furthermore, we combine the modified training method mentioned in Section \ref{sec5.1}, and the robustness of class 3 becomes 43.5\% (Line 6), which means that the robustness of the most vulnerable class in CIFAR-10 has been greatly improved.

\section{Conclusion} \label{sec6}
In this paper, we have a closer look at the class-wise properties of the robust model based on the observation that robustness between each class has a recognizable gap.
We conduct systematic analysis and find: 
1) In each dataset, classes can be divided into several subgroups, and intra-group classes are easily flipped by each other. 
2) The emergence of the unbalanced robustness is closely related to the intrinsic properties of the datasets.
Furthermore, we make full use of the properties of the vulnerable classes to propose an attack that can effectively reduce the robustness of these classes, thereby increasing the disparity among classes.
Finally, in order to alleviate the robustness difference between classes, we propose two methods to improve the robustness of the most vulnerable class in CIFAR-10 (\textit{i.e.}, class 3): 
1) At the training phase: Modify loss function according to group-based relation between classes.
2) At the inference phase: Change the background of the original images.
We believe our work can contribute to a more comprehensive understanding of adversarial training and let researchers realize that the class-wise properties are crucial to robust models.

\section{Acknowledgements}
This work is supported in part by Key R\&D Projects of the Ministry of Science and Technology (No. 2020YFC0832500), National Natural Science Foundation of China (No. 61625107, No. 62006207), National Key Research and Development Program of China (No. 2018AAA0101900), the Fundamental Research Funds for the Central Universities and Zhejiang Province Natural Science Foundation (No. LQ21F020020). Yisen Wang is partially supported by the National Natural Science Foundation of China under Grant 62006153, and CCF-Baidu Open Fund (OF2020002).
\bibliographystyle{ACM-Reference-Format}
\bibliography{sample-base}

\newpage
\appendix
\section{Appendix: hyperparameters for reproducibility}
\subsection{Hyperparameters for defenses in Section \ref{sec4}} \label{A1}
\; \; \textbf{MNIST setup.}\, Following \citet{zhang2019theoretically}, we use a four-layers CNN as the backbone. In the training phase, we adopt the SGD optimizer with momentum $0.9$, weight decay $2 \times 10^{-4}$ and an initial learning rate of $0.01$, which is divided by 10 at the 55$^{{\rm th}}$, 75$^{{\rm th}}$ and 90$^{{\rm th}}$ epoch (100 epochs in total). Both the training and testing attacker are 40-step PGD (${\rm PGD^{40}}$) with random start, maximum perturbation $\epsilon=0.3$ and step size $\alpha=0.01$. 

\textbf{CIFAR-10 \& CIFAR-100 setup.}\, Like \citet{wang2019improving} and \citet{zhang2019theoretically}, we use ResNet-18 \citep{he2016deep} as the backbone. In the training phase, we use the SGD optimizer with momentum $0.9$, weight decay $2 \times 10^{-4}$ and an initial learning rate of $0.1$, which is divided by 10 at the 75$^{{\rm th}}$ and 90$^{{\rm th}}$ epoch (100 epochs in total). The training and testing attackers are ${\rm PGD^{10}}/{\rm PGD^{20}}$ with random start, maximum perturbation $\epsilon=0.031$ and step size $\alpha=0.007$. 

\textbf{SVHN \& STL-10 setup.}\, All settings are the same to CIFAR-10 \& CIFAR-100, except that the initial learning rate is 0.01.

\textbf{ImageNet setup.}\, Following \citet{wong2019fast}, we use ResNet-50 \citep{he2016deep} as the backbone. 
Specifically, in the training phase, we use the SGD optimizer with momentum $0.9$ and weight decay $2 \times 10^{-4}$. 
A three-stage learning rate schedule is used as the same with \citet{wong2019fast}.
The training attacker is ${\rm FGSM}$ \citep{goodfellow2014explaining} with random start, maximum perturbation $\epsilon=0.007$, and
the testing attacker is ${\rm PGD^{50}}$ with random start, maximum perturbation $\epsilon=0.007$ and step size $\alpha=0.003$. 

\subsection{Hyperparameters for attacks in Section \ref{sec4.4}} \label{A2}
\; \; \textbf{FGSM setup.} Random start, maximum perturbation $\epsilon=0.031$.

\textbf{PGD} \textbf{setup.} Random start, maximum perturbation $\epsilon=0.031$. For RST model, step size $\epsilon=0.01$ and steps $\alpha=40$, following \citet{carmon2019unlabeled}. For other models, step size $\epsilon=0.003$ and steps $\alpha=20$.

\textbf{CW}$\mathbf{_{\infty}}$ \textbf{setup.}
Binary search steps $b=5$, maximum perturbation times $n=1000$, learning rate $lr=0.005$, initial constant $c_0=0.01$, $\tau$ decrease factor $\gamma=0.9$.
Similar to \citet{carmon2019unlabeled}, we randomly sample 2000 images to evaluate model robustness, and 200 images per class.

\textbf{Transfer-based attack setup.} All settings are the same to PGD for the substitute standard model.

$\bf {\mathcal{N}}$ \textbf{attack setup.} Random start,  maximum perturbation $\epsilon=0.031$, population size $n_{pop} = 300$, noise standard deviation $\sigma = 0.1$ and learning rate $lr = 0.02$. Similar to \citet{li2019nattack}, we randomly sample 2000 images to evaluate model robustness, and 200 images per class.

\section{Appendix: Discussion with the class-wise accuracy in standard training} \label{A3}
\begin{table}[t]
	\centering
	\tiny
	\caption{Statistical metric: class-wise variance(CV) and maximum class-wise discrepancy(MCD) of the benchmark datasets}
	\begin{tabular}{ccccc}
		\hline
		\multirow{2}{*}{Dataset} & \multicolumn{2}{c}{\begin{tabular}[c]{@{}c@{}}Standard training\\ (natural accuracy)\end{tabular}} & \multicolumn{2}{c}{\begin{tabular}[c]{@{}c@{}}Adversarial training\\ (robustness accuracy)\end{tabular}} \\ \cline{2-5} 
		& CV(\%)                                           & MCD(\%)                                          & CV(\%)                                              & MCD(\%)                                             \\ \hline \hline
		MNIST                    & 0.11                                            & 0.98                                             & 2.71                                               & 6.00                                                \\
		SVHN                     & 2.78                                            & 5.47                                             & 110.25                                             & 37.10                                               \\
		CIFAR-10                 & 10.75                                           & 10.80                                            & 284.43                                             & 52.80                                               \\
		STL-10                   & 151.98                                          & 36.13                                            & 250.06                                             & 51.00                                               \\
		CIFAR-100                & 115.62                                          & 30.00                                            & 212.82                                             & 44.00                                               \\
		ImageNet                 & 501.70                                          & 68.00                                            & 637.21                                             & 78.00                                               \\ \hline
	\end{tabular}
	\label{table1}
\end{table}

The body part of this paper mainly focuses on the adversarial robustness of the model. One possible concern is that even with standard training, the class-wise accuracy may not be exactly the same. Indeed, in order to highlight the difference of this phenomenon between adversarial training and standard training.
We also report the CV and MCD of these two training methods since these metrics can quantitatively reflect the discrepancy between classes.

In our experiment, the pipeline and hyperparameters of the standard training are consistent with the adversarial training, except that the adversarial examples are not added to the training set.
From the results in Table \ref{table1}, it can be found that in all datasets, the class-wise variance (CV) and maximum class-wise discrepancy (MCD) of the adversarial training are significantly larger than the standard training, especially for the most commonly used dataset in the adversarial community, CIFAR-10, the CV of adversarial training is 28 times that of standard training, and the MCD of adversarial training is 5 times that of standard training.
This shows that class-wise properties in the robustness model are worthy of attention.

\end{document}